\documentclass{article}
\usepackage{log_2022}           

\usepackage{booktabs}            % professional-quality tables
\usepackage{multirow}            % tabular cells spanning multiple rows
\usepackage{amsfonts}            % blackboard math symbols
\usepackage{graphicx}            % figures
\usepackage{duckuments}          % sample images
\usepackage{longtable}
\usepackage{subfigure}
\usepackage[numbers,compress,sort]{natbib}
\title[Estimating fire Duration using regression methods]{Estimating Wildfire Duration using regression methods}
\author[Hansong Xiao]{%
Hansong Xiao\\
\institute{Imperial College London}\\
\email{hx221@ic.ac.uk}
}

\begin{document}

\maketitle
% ****************************************Abstract**************************************************
\begin{abstract}
Wildfire forecasting problems usually rely on complex grid-based mathematical models, mostly involving Computational fluid dynamics(CFD) and Celluar Automata, but these methods have always been computationally expensive and difficult to deliver a fast decision pattern. In this paper, we provide machine learning based approaches that solve the problem of high computational effort and time consumption. This paper predicts the burning duration of a known wildfire by RF(random forest), KNN, and XGBoost regression models and also image-based, like CNN and Encoder. Model inputs are based on the map of landscape features provided by satellites and the corresponding historical fire data in this area. This model is trained by happened fire data and landform feature maps and tested with the most recent real value in the same area. By processing the input differently to obtain the optimal outcome, the system is able to make fast and relatively accurate future predictions based on landscape images of known fires.
\end{abstract}

% ****************************************Introduction**************************************************
\section{Introduction}

As humans evolved, fire seems to become more controllable. But wildfires are not included, wildfires are a natural phenomenon that is difficult to predict and occur with high frequency, more than 4000 fires happened in CA, USA from 2003 to 2016 \citep{altas_fire}. More seriously, wildfire can even be called a natural hazard \citep{1}. Wildfires endanger most of the world\citep{2}, 4 million $km^2$ to 6 million $km^2$ (depends on different estimate methods) of land are affected by wildfire every year\citep{3}, which is almost half the area of Europe. 

Wildfire has a huge negative impact on earth's ecosystem, risking natural systems and health, hydrogeomorphic effects from wildfire jeopardize the quality of daily water in both human society and the ecosystem\citep{4}. Wildfire also threatens the world economy, house price of northwest Montana, a potential resort area in the USA, was depreciated by unpredictable long-term wildfire\citep{5}. It also results in serious injury or death for animals, a huge group of sheep burned in a wildfire in 2006\citep{6}. To prevent multiple hazards, researchers have begun to engage in developing wildfire prediction. Cellular Automata \citep{7} and CFD (Computational Fluid Dynamic) \citep{8} were widely used. These traditional methods are normally based on building a complex mathematical model that takes all variables and interactions between features into consideration. One of the most important and pioneering models is Rothermal's equation\citep{rothermel1972mathematical}, but most of these models are hard to understand and lack of interpretability. Basically, a solution to the grid-based method is solving dynamic function or a system of partial differential equation\citep{cellular_automate}, which is notoriously time-consuming. Thus, grid-based methods were intensive in computation and required resources that are nearly impossible to achieve a fast-decision pattern. With the development of artificial intelligence, machine learning becomes a new research direction\cite{mlhistory,cheng2022data,cheng2022parameter} on wildfire prediction. Satellite imagery was used to help analyze geomorphic features to build models for better forecasting \cite{zhong2023reduced,cheng2023generative}. However, it is challenging to develop a general and effective ML model in a worldwide range for forecasting wildfire, due to the complicity of environmental parameters in different areas. 

Currently, most ML models put insight into a specific direction, like the occurrence of human-caused wildfire\citep{human-caused_wildfire}, forecasting sensitivity map of wildfires\citep{lever2023social}, and predicting the likelihood of wildfires \citep{probabilityofwildfire}. Most of the current research is focusing on possible future wildfires, which requires management groups put energy in every possible location. 80 percent of fires last no more than one day based on our extracted dataset \citep{altas_fire}, indicating fire management groups are unable to respond effectively in a limited time. It would be wise to keep eye on preventing potential large fires. In this paper, we aim to deliver a fast-decision model that focuses on predicting the likely extent of ongoing fires. ML models were split into two parts, simple regression models and image-based models. In the ML regression part, three methods were applied, including RF\citep{random_forest}, KNN\cite{kNN,gong2022efficient}, and XGBoost\citep{chen2016xgboost} to predict fire duration. In the image-based part, two CNN-based models were introduced, including multi-layer CNN and Encoder-Based CNN \cite{cheng2023generalised}.

% ****************************************Data Curation**************************************************
\section{Data curation}

\subsection{Data Gathering}
\begin{figure}[htbp]
    \centering
    \includegraphics[width = 13cm, height = 8cm]{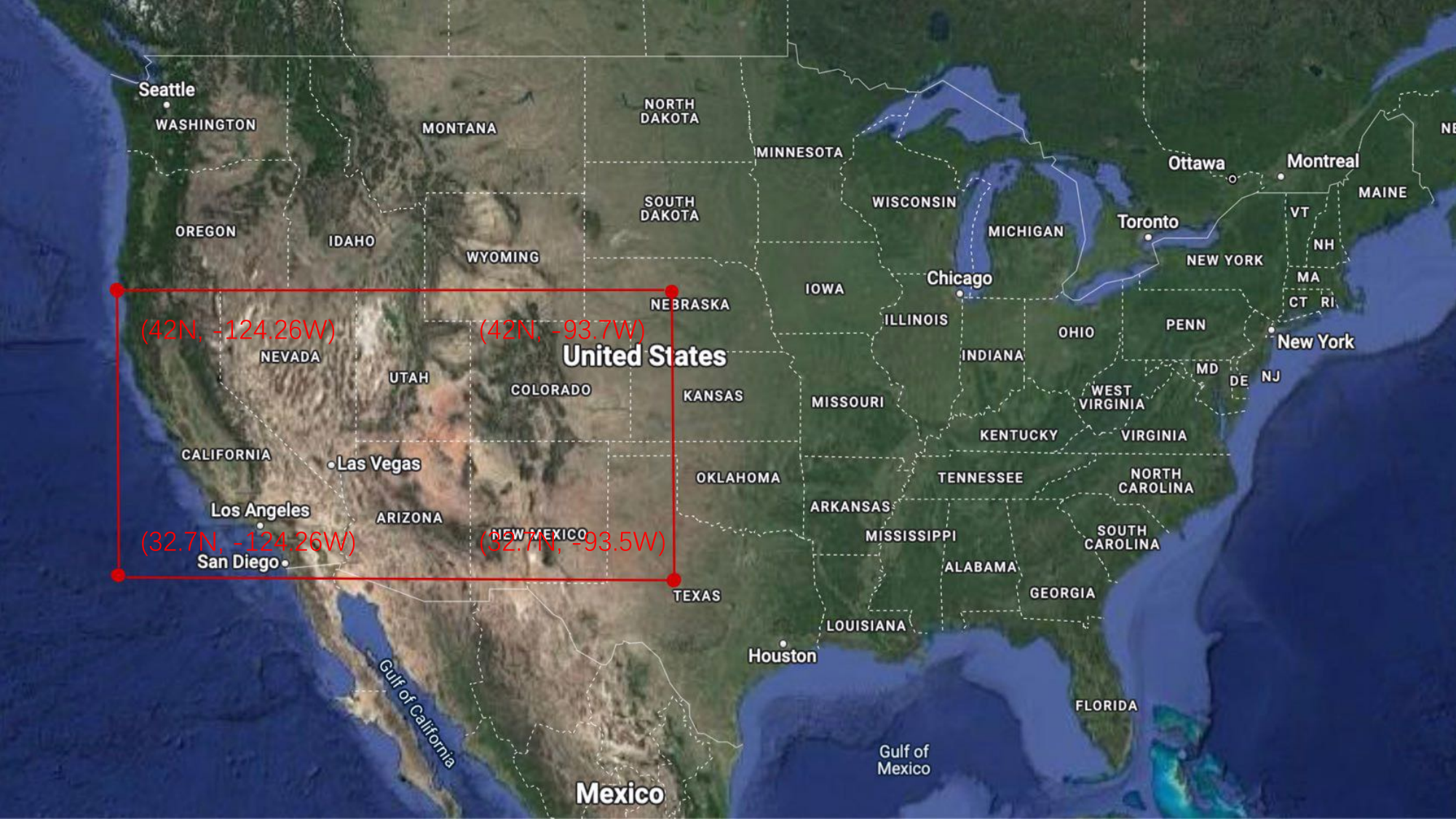}
    \caption{Target area}
    \label{Target Area}
\end{figure}

We train and validate our model using most of the fires that occurred in 13 years in a specific region, as shown in figure \ref{Target Area}, ranging from 32.7N to 42N, -124.26W to -93.5W, which is mainly the Midwestern region of the United States, including California, Phoenix, Utah, Colorado, New Mexico, Kansas, and Texas. This part was chosen due to the similarity of geomorphology among them \citep{geofeature}, the increasing number of wildfires with an expanded rate of a burned area at 355$km^2$ per year \citep{firetrend}, and an enormous economic loss excesses billions of dollars per year \citep{ecoloss}. If we consider the entire U.S., the geomorphic disparity between the West Coast and the East and the sparsity of wildfire on the East would impact the accuracy of the model. 

We decide to use \textit{altas fire dataset}\citep{altas_fire} for fire data, which includes the coordinates, start and end times, burned area, perimeter, and duration of the fires. We selected fire duration as the label value, considering the duration becomes more significant in evaluation wildfire\citep{duration}. Eighty percent of fires are small, and last less than one day. To avoid data imbalance, we randomly dropped half of the small fire (duration less than 1 day).

\begin{figure}[htbp]
    \centering
    \includegraphics[width = 13cm, height = 8cm]{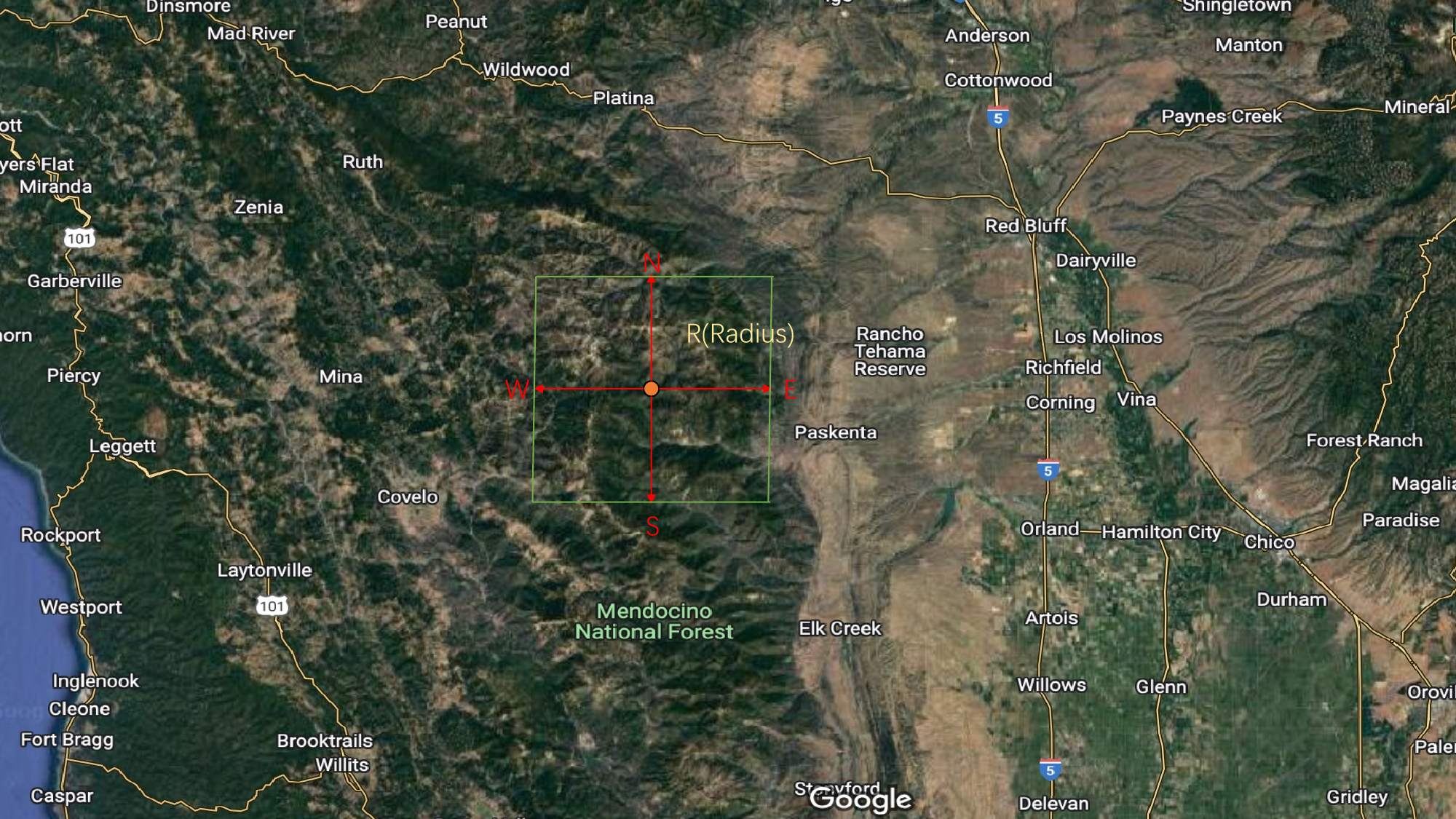}
    \caption{Example of fire extension}
    \label{expand map}
\end{figure}

The database was split into two parts, from 2003 to 2013 as a train set, and from 2013 to 2016 as a test set. The ratio of fire number between the train and test is around 0.75, which is a reasonable number for data splitting. We expand each target coordinate in four directions for 30 kilometers range which is demonstrated in figure \ref{expand map}. The yellow point represents the coordinate of a random fire, and each red line shows how the 30 kilometers radius of the extension area. the square surrounded by green lines is the feature map to gather. We use a 30km radius size to avoid repeat data extraction work, which is time-consuming if a smaller radius were taken into consideration. Currently, three global scale databases were involved in this experiment, density map of above(or below) ground living biomass storage \citep{biomass}, landform slope map \citep{slope}, land cover map of multiple surface characteristics (e.g. grass coverage, tree coverage, snow converge) \citep{landcover}, and climate data corresponding to specific location range from 1979 to 2020 (e.g. precipitation, wind speed) \citep{precipitation}. In general, five input variables, tree cover, $^r{\mathbf{X}}^{tc}_{z} = \{ ^rx^{tc}_{i,j}\}$, geo-slope, $^r{\mathbf{X}}^{slope}_{z} = \{ ^rx^{slope}_{i,j}\}$, grass cover, $^r{\mathbf{X}}^{gc}_{z} = \{ ^rx^{gc}_{i,j}\}$, wind in u direction, $^r{\mathbf{X}}^{windu}_{z} = \{ ^rx^{windu}_{i,j}\}$ and wind in v direction $^r{\mathbf{X}}^{windv}_{z} = \{ ^rx^{windv}_{i,j}\}$ were obtained to predicting $^r{\mathbf{D}}^z$, duration. Here, $i, j$, the size of the matrix, for slope range from 0 to 200, grass cover and tree cover range from 0 to 700, $z$, and wind range from 0 to 4, fire index, $r$, the radius of extension. A detailed table of the dataset used is shown in table \ref{table1}.

\begin{table}[h]
\resizebox{\textwidth}{20mm}{
\begin{tabular}{|c|c|c|c|c|c|}
\hline
\textbf{Dataset}         & \textbf{Resolution}  & \textbf{Sources}                            & \textbf{Image Size} \\ \hline
\begin{tabular}[c]{@{}l@{}}Global Fire Atlas with Characteristics \\ of Individual Fires\end{tabular} & None & MODIS\citep{altas_fire} & None\\ \hline
\begin{tabular}[c]{@{}l@{}} Global ALOS mTPI\newline   \end{tabular}  & 270 Metres   & ERGo\citep{slope}   & (200, 200)     \\ \hline
\begin{tabular}[c]{@{}l@{}}opernicus Global Land \\ Cover Layers \end{tabular}   & 100 Metres   & \begin{tabular}[c]{@{}l@{}}GlobCover, LC-CCI, \\ MODIS, PROBA-V \end{tabular}\citep{landcover}  & (700, 700)     \\ \hline
\begin{tabular}[c]{@{}l@{}} ERA5 Monthly Aggregates\\ - Latest Climate Reanalysis \end{tabular} & 27830 Meters & ECMWF\citep{precipitation}  & (4, 4)        \\ \hline
\end{tabular}}
\caption{Dataset Detail}
\label{table1}
\end{table}

% \newpage

\subsection{Data Standardization}
Data standardization was applied for all inputs due to different scales of vegetation cover, normalizing all inputs in the same scale. \textit{min-max} normalization \citep{minmax}were adopted to resize all data into 0 to 1 range, as the equation is shown below, where $X$ indicates each input matrix, where $f$ is the feature name, $r$ representing radius, and $z$ regarding fire index. 

\begin{equation}
    ^rX^{f}_z = \frac{^rX^{f}_z-min(^rX^{f}_z)}{max(^rX^{f}_z)-min(^rX^{f}_z)}
\end{equation}

% ****************************************Methodology**************************************************

\section{Methodology}
This section discusses several models designed for fast-decision fire prediction based on different algorithms. To ensure consistency across the results, the input and output for models constructed on different bases are standardized. Three ML regression models are employed, maintaining the same input. Besides, the image-based ML model is used for comparison with the same input, input contains five possible different images, and five feature maps, which are grass cover, tree cover, landform slope, and two wind components(windu and windv). The output for these models is fire duration. We make use of three datasets, data from 2003 to 2013 in the target area regarded as train data, from 2013 to 2016 classified as test data.

The prediction result is hard to evaluate by Root Mean Square Error(RMSE) because the severity range of fire is not precise. In the numerical aspect, it would be a prediction with more than 50 percent error if a test data has a predicted value of 20 but with a label value of 30. However, in the aspect of fire prevention, both 20 and 30 indicate this fire exists for a long term, showing the seriousness of this fire, thus it could be defined as a reasonable prediction, which can be classified as the true prediction. To balance the ambiguous prediction result from numerical and prevention aspects, another measurement was introduced to evaluate all the possible outcomes, which is called an accuracy plot. 

A prediction was regarded as a True Prediction(TP) if the prediction is greater than the critical value, otherwise, it is a False Prediction (FP). Calculating the ratio of positive out of total prediction, as the equation shown below equation \ref{equation acc}, gives the accuracy of prediction. This ratio varies with different critical values, critical values with a range of 0 to the maximum label value. Thus, an accuracy plot was obtained and the lowest point in the accuracy curve presented the worst situation of each method. 
\begin{equation}
    Accuracy = \frac{TP}{TP+FP}
    \label{equation acc}
\end{equation}
where TP is the true prediction, which has a prediction greater than the critical value, FP is false prediction, which has a prediction smaller than the critical value. 

Meanwhile, another equation was used to compute RMSE of each single prediction, which gives the ratio between predicted days and the original fire duration. 

\begin{equation}
    RMSE = \sqrt{\frac{1}{n} \sum_{i}^{n} (\frac{\left\| ^rD^z_i - \hat{^rD^z_i}\right\|}{\left\|\hat{^rD^z_i}\right\|})}
\end{equation}

where $^rD^z_i$ is the label value in radius $r$ and fire index $z$, $\hat{y}$ is the prediction value.

\subsection{Regression modeling}
The Regression model aims to give a simple prediction. This model regards different standardized features (vegetation cover, slope of topographic map, wind, and precipitation)as input to predict fire severity. Output is a measurement variable that could represent the fire severity (fire size, fire duration), fire duration was used because it shows more accuracy rather than other label value. For the radius, 5km was adopted here as the figure \ref{trend_map} shows that radius has nearly no effect on the result.

\subsubsection{Data preprocessing}
We initialized the collected data and divided it into different ranges of inputs.
In order to explore the affection of different ranges of input, we cropped the original image with a radius of 30km to a smaller radius, from the center point to the corresponding scale, and calculate the average value of each image in a different radius. Then, we obtained a series of average values with different radius. The average value is the input for the regression model. For example, the average value of 80 $\times$ 80 in the middle of the original map will be used as input if we want a range of 3km.

\begin{equation}
    ^nX_{f}^{r} = \frac{\sum{x_{ij}}}{i\times j}
\end{equation}
where X is the average value of the input, r is the radius, n is the fire index and f is the feature name. Below is a demonstration of the crop process when the radius is equal to three kilometers.

$$
X=
\begin{bmatrix}
x_{1,1} & \cdots   & x_{1,600}   \\
\vdots  & \ddots   & \vdots  \\
x_{600,1}  & \cdots\  & x_{600,600}  \\
\end{bmatrix}
\stackrel{radius = 3km}{\longrightarrow}
\begin{bmatrix}
x_{270,270}  & \cdots   & x_{270,330}   \\
\vdots  & \ddots   & \vdots  \\
x_{330, 270}  & \cdots\  & x_{330,330}  \\
\end{bmatrix}
$$

\subsubsection{Random Forest regression model}
We use random forest (RF) \citep{random_forest} as the first regression method. 
Random forest is an extension branch of the bagging algorithm, and RF is based on a series of Decision Trees(DTs). Random Forest develops models, which consist of a series of trees, that predict individually and separately. The output integrates the results of each predictor by a majority rule to forecast an output or one classification. In the regression study, different sets of samples were obtained from an initial dataset, the result is the average value of all the trees. The hyperparameters contain the number of individual predictors \textit{n}, and the depth of each predictor \textit{d}. 

\subsubsection{K-nearest neighbors algorithm model}
K-nearest neighbors (kNN) is a supervised learning algorithm, working on both classifier and regression \citep{kNN}. In classification, KNN assigns each single data point into the same category as the highest frequency of \textit{k} nearest neighbors. In regression, calculating the average label value of \textit{k} nearest point of the input train data to get the prediction value. In KNN model, input is the mean value of each feature map, and we keep using fire duration as output. The hyperparameter is the number of neighbors \textit{k}. 

% graph may need

\subsubsection{XGBoost}
Extreme Gradient Boosting \citep{chen2016xgboost} is another ensemble learning algorithm derived from  Gradient Boost Decision Tree (GBDT) \citep{ke2017lightgbm}. XGBoost takes every input data into account as each input results in one tree. XGBoost also considers the gradient of data, which saves calculation time. XGBoost gives regularized term for the loss function, which penalties the complexity of the model and compensates the learning term, thus XGBoost is more effective and more capable of preventing over-fitting. The researched hyperparameters are the number of estimator \textit{n}, learning rate \textit{l}, and max depth \textit{d}.

\subsection{Image Based Modelling}
This part aims to explore the performance of the image-based model in predicting fire severity. To ensure the accuracy of forecasting, we tried permutation of different inputs and decide vegetation cover and landform slope are the final input as they give the best accuracy. The same output value or label value, fire duration, was selected to ensure the comparability between models. We make use of the image as input instead of the average value. All feature maps were resized to the same and placed in different channels of one data array as the traditional CNN. Moreover, the fire duration dataset from 2003 to 2013 was used as a train dataset and from 2013 to 2016 as a test dataset. A 5km data radius was selected to avoid unnecessary computation and keep the same radius as the regression ML model.

\subsubsection{Data preprocessing}
Different ranges of images are not interested in this model, since the radius presents little impact on final result. The data preprocessing method applied is resized, which scale all image to the same size by an interpolation method. Resize keeps the image the same but only change width and height. This approach calculates the value of a new image pixel based on the value of the original pixel points by bi-linear interpolation\citep{bi-linear} shown below. 

\begin{equation}
    x_{ij} = [(1-p)x_{ij} + px_{i+1, j}](1-q) + [(1-p)x_{i, j+1} + px_{i+1, j+1}]q
\end{equation}
where p, q is the fraction of distance between points, x is the pixel point, i, j is the index of the input point.

\subsubsection{Multilayer Convolutional Neural Network (CNN)}

\begin{figure}[h]\centering
    \centering
    \includegraphics[width = 13cm, height = 5cm]{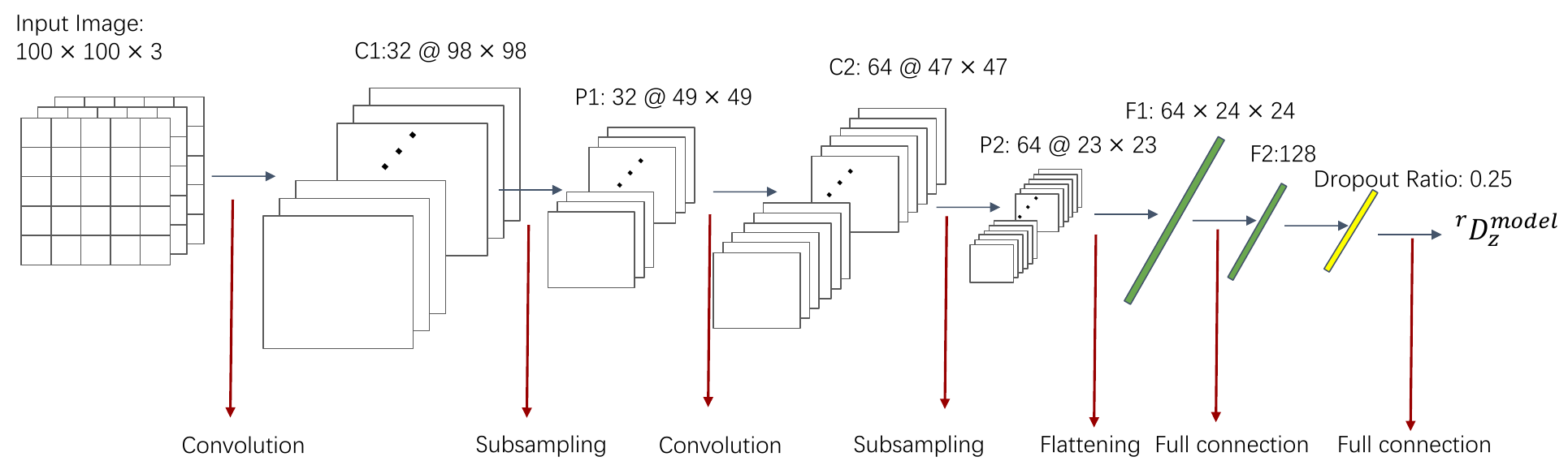}
    \caption{CNN Structure}
    \label{fig3}
\end{figure}

CNN, one of the most famous machines learning neural network\citep{CNN}, extracted the significance feature from original data by mathematics operation convolution, which is used in this model for wildfire prediction. CNN has the ability to process large amounts of data due to the number of its neural, CNN could better recognize and learn for image type input. The structure of this model is shown in figure \ref{fig3}. ReLu and linear was used as activation function. A $3 \times 3$ kernel size and 32 channel numbers are selected in the convolutional layer to maximum keep the feature of the image. Two dropout layer is added in this model in case of over-fitting. To avoid gradient vanishing \citep{leakyrelu}, we added a LeakyRelu layer after the dense layer. Detailed structures on this model are shown on table \ref{table_cnn}. For the hyperparameters, 100 epochs, 128 batch size is adopted and $adam$ optimizer is used. 

\begin{table}[h]
\centering
\caption{Structure of Multilayer CNN}
\label{table_cnn}
\begin{tabular}{@{}ccc@{}}
\toprule
Layer Type   & Output Shape   & Activation \\ \midrule
Input        & (100, 100, 5)  &            \\
Conv 2D(3$\times$3) & (98, 98, 32) & ReLu       \\ 
MaxPooling 2D(2$\times$2) & (49, 49, 32) &\\
Conv 2D(3$\times$3) & (47, 47, 64) & ReLu \\
Dropout(0.25)           & (47, 47, 64) &\\
MaxPooling 2D(2$\times$2) & (23, 23, 64) &\\
Flatten                 & 36856  &\\
Dense               & 128       & linear \\
LeakyRelu          & 128       & LeakyReLu(0.2)\\
Dropout(0.25)       & 128       &\\
Dense               & 1         & linear \\
\bottomrule
\end{tabular}
\end{table}

\subsubsection{Encoder Based Convolutional Neural Network (CNN)}
Autoencoder \citep{sparseautoencoder} is an unsupervised machine learning technique to extract key features from input and ignore inessential points, learning the significant sets of data of input. Autoencoder consists of two parts, encoder and decoder, encoder responds to learning the crucial point from initial input, decoder is used to reduce refined features and return to its original state. In this model, we make use of an encoder, which is shown in table \ref{table_encoder_cnn}, to perform feature extraction. The encoder is trained jointly with an ML structure as shown in table \ref{table_encoder_cnn}. Two different encoders were applied in this model, due to the different sizes of the input image, vegetation map with a size of 100$\times$100, and landform slope only has 30$\times$30. Two encoders extracted features from the input separately and trained the concatenated dense layer. The detailed structure map is shown in figure \ref{encoder_based_cnn} and table \ref{table_encoder_cnn}. In this study, the batch size, epochs, and optimizer were set to 128, 20, and $adam$, respectively. 

\begin{longtable}{@{}ccc@{}}
\caption{Structure of Encoder Based CNN}\label{table_encoder_cnn}\\
\toprule
Layer Type   & Output Shape   & Activation \\ \midrule
\textbf{EncoderA}       &       & \\
Input        & (100, 100, 1)  &            \\
Conv 2D & (100, 100, 32) & ReLu       \\ 
MaxPooling 2D(2$\times$2) & (50, 50, 32) &\\
Conv 2D & (50, 50, 16) & ReLu \\
MaxPooling 2D(2$\times$2) & (25, 25, 16) &\\
Conv 2D & (25, 25, 16) & ReLu \\
MaxPooling 2D(2$\times$2) & (13, 13, 16) &\\
Flatten                 & 2704  &\\
LeakyRelu          & 2704       & LeakyReLu(0.3)\\
Dense               & 169       & linear \\
LeakyRelu          & 169       & LeakyReLu(0.2)\\
Dense               & 20         & linear \\
\textbf{EncoderB}&  &   \\
Input        & (30, 30, 1)  &            \\
Conv 2D & (30, 30, 32) & ReLu       \\ 
MaxPooling 2D(2$\times$2) & (15, 15, 32) &\\
Conv 2D & (15, 15, 16) & ReLu \\
MaxPooling 2D(2$\times$2) & (8, 8, 16) &\\
Flatten                 & 1024  &\\
LeakyRelu          & 1024       & LeakyReLu(0.3)\\
Dense               & 64       & linear \\
LeakyRelu          & 64       & LeakyReLu(0.2)\\
Dense               & 10         & linear \\
\textbf{Encoder Based CNN}&  &   \\
Input A                & (100, 100, 1)  &            \\
Input B                & (30, 30, 1)  &            \\
Encoder A (Input A)      & 20             & \\
Encoder B (Input B)     & 10        & \\
Concatenate             & 50        & \\
Dense                   & 50        & ReLu \\ 
Dense                   & 50        & Linear \\
\bottomrule
\end{longtable}

\subsubsection{Hyperparameter Tuning}
To filter the optimal parameter value, grid search was applied to all models and compared the resulting loss on the test dataset. The number of estimators and max depth were studied for random forests. One more parameter, learning rate, was researched for XGBoost. Batch Size, epochs, and optimizer were tested for both image-based methods. Detailed results are shown in table \ref{grid search}. 

\begin{table}[h]\centering
\caption{Optimal Parameter of Grid Search}
\label{grid search}
\begin{tabular}{@{}ccc@{}}
\toprule
Hyperparameters  & Grid Search Space   & Optimal Set \\ \midrule
\textbf{Random Forest} & & \\
Number of estimator & {5, 10, 20, 50, 100} &    50\\
Max depth &   {1, 5, 10, 20, 50}  &    5\\
\textbf{KNN} &  &\\
Number of neighbors \textit{k} &  {5, 10, 20, 50}  & 10 \\
\textbf{XGBoost} & &\\
Number of estimator &   {5, 10, 20, 50} & 50\\
Learning Rate &     {0.05, 0.1, 0.02} &     0.1 \\
Max depth   &   {1, 5, 10, 20, 50}  &   5 \\
\textbf{Mutil-layer CNN} &  & \\
Batch Size  & {32, 64, 128, 256}  &  128\\
Epochs     & {10, 20, 50, 100}     & 50\\
Optimizer &     {SGD, Adam}     & Adam \\
\textbf{Encoder-based CNN} & &\\
Batch Size  & {32, 64, 128, 256}  &  128\\
Epochs     & {10, 20, 50, 100}     & 50\\
Optimizer &     {SGD, Adam}     & Adam \\
\bottomrule
\end{tabular}
\end{table}

% ************************************Result & Disscusion***********************************************
% \newpage
\section{Results and Discussion}

\begin{figure}[htbp]
\centering
\subfigure[Prediction Plot]{
\begin{minipage}[ht]{0.45\textwidth}
\centering
\includegraphics[width = 6 cm, height = 6 cm]{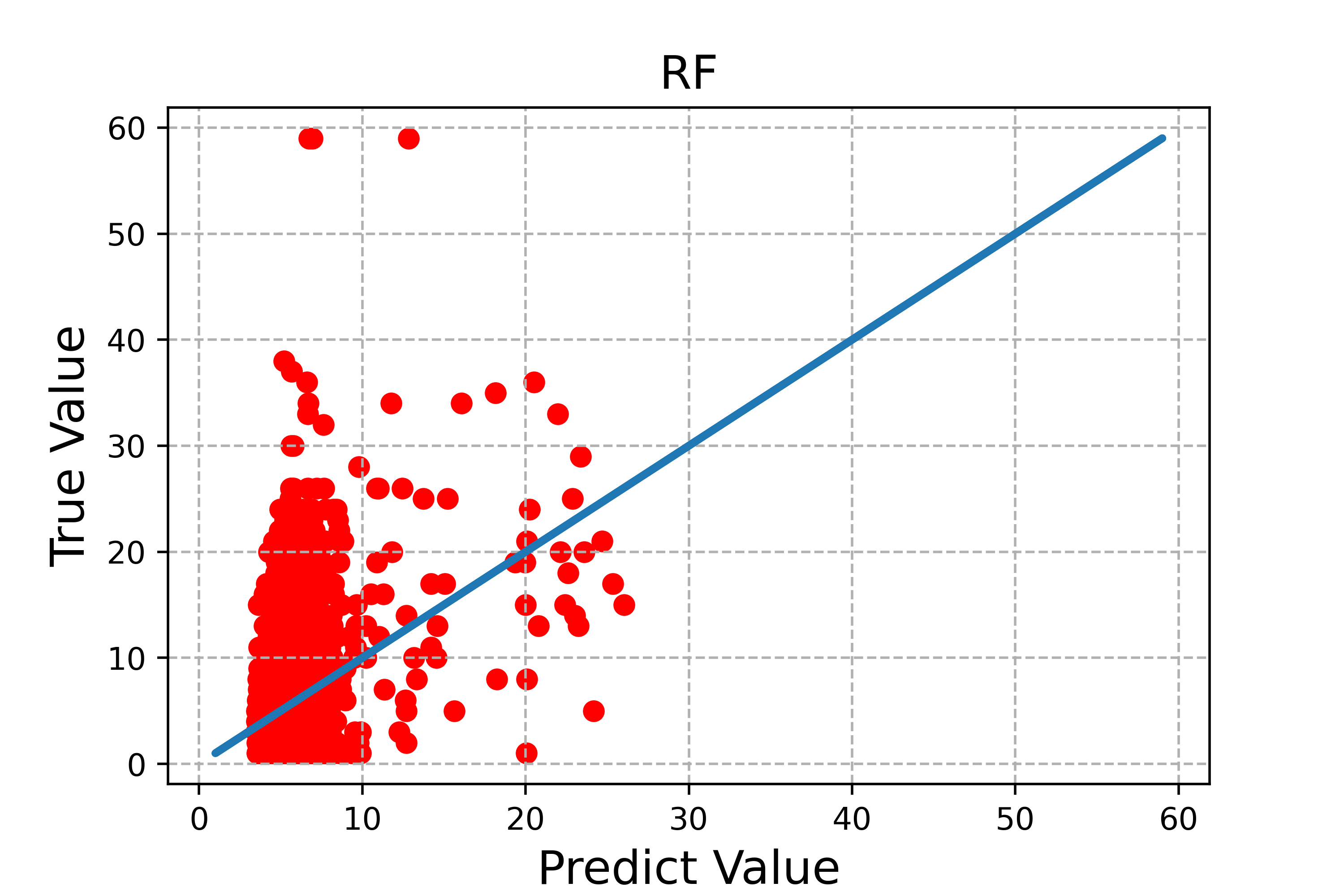}
\includegraphics[width = 6 cm, height = 6 cm]{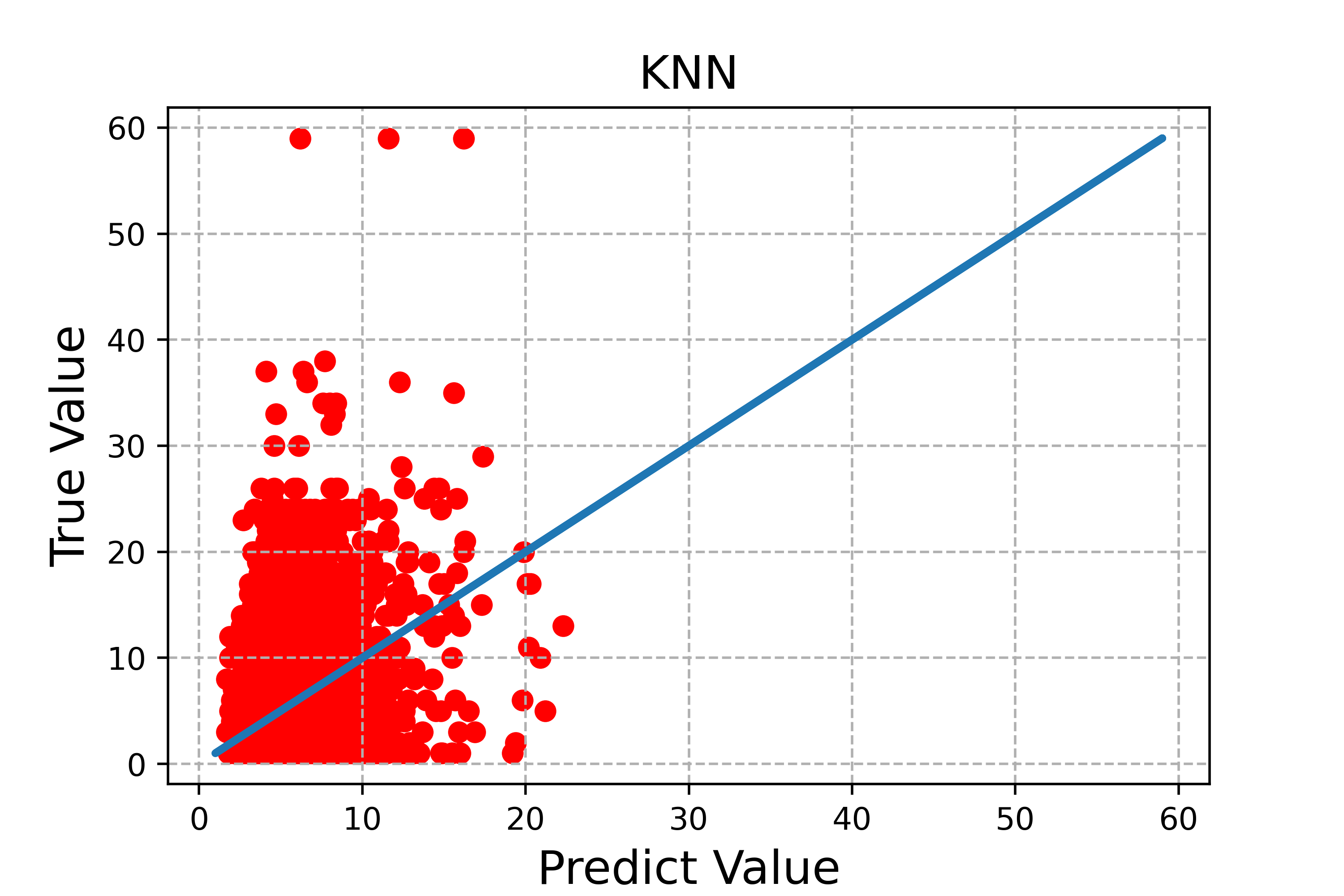}
\includegraphics[width = 6 cm, height = 6 cm]{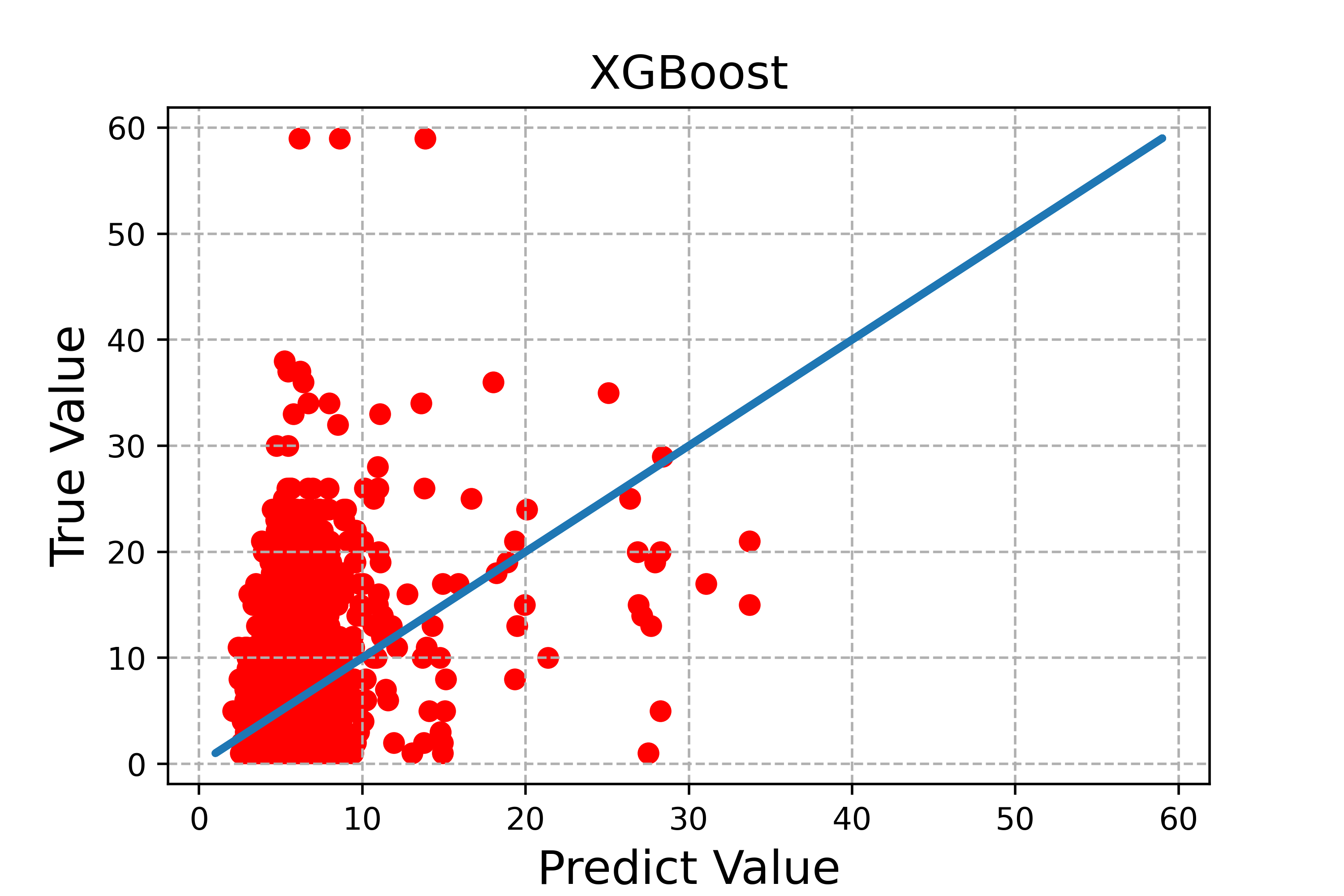}
%\caption{fig1}
\end{minipage}
\label{regression prediction}
}%
\subfigure[Accuracy Plot]{
\begin{minipage}[ht]{0.45\textwidth}
\centering
\includegraphics[width = 6 cm, height = 6 cm]{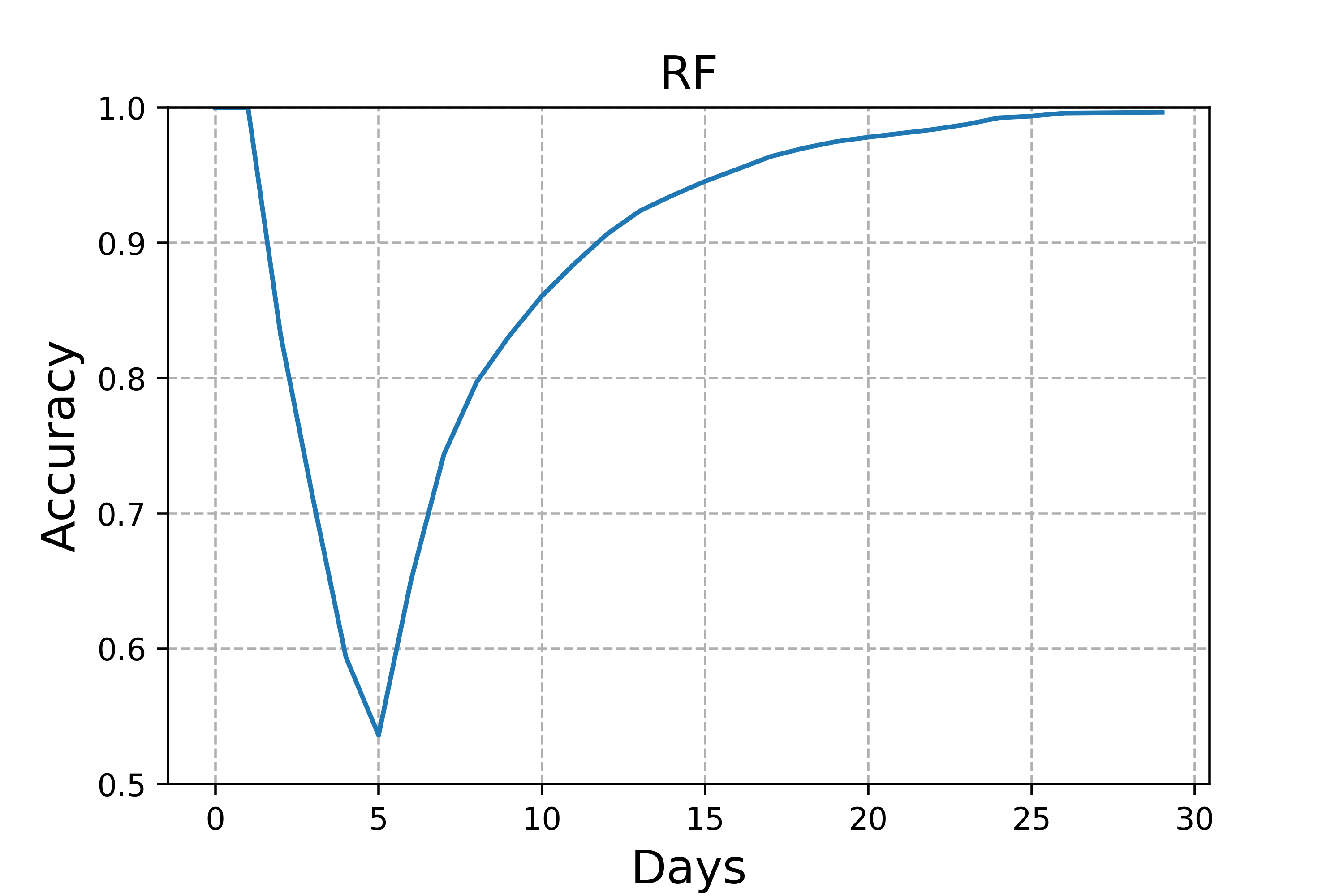}
\includegraphics[width = 6 cm, height = 6 cm]{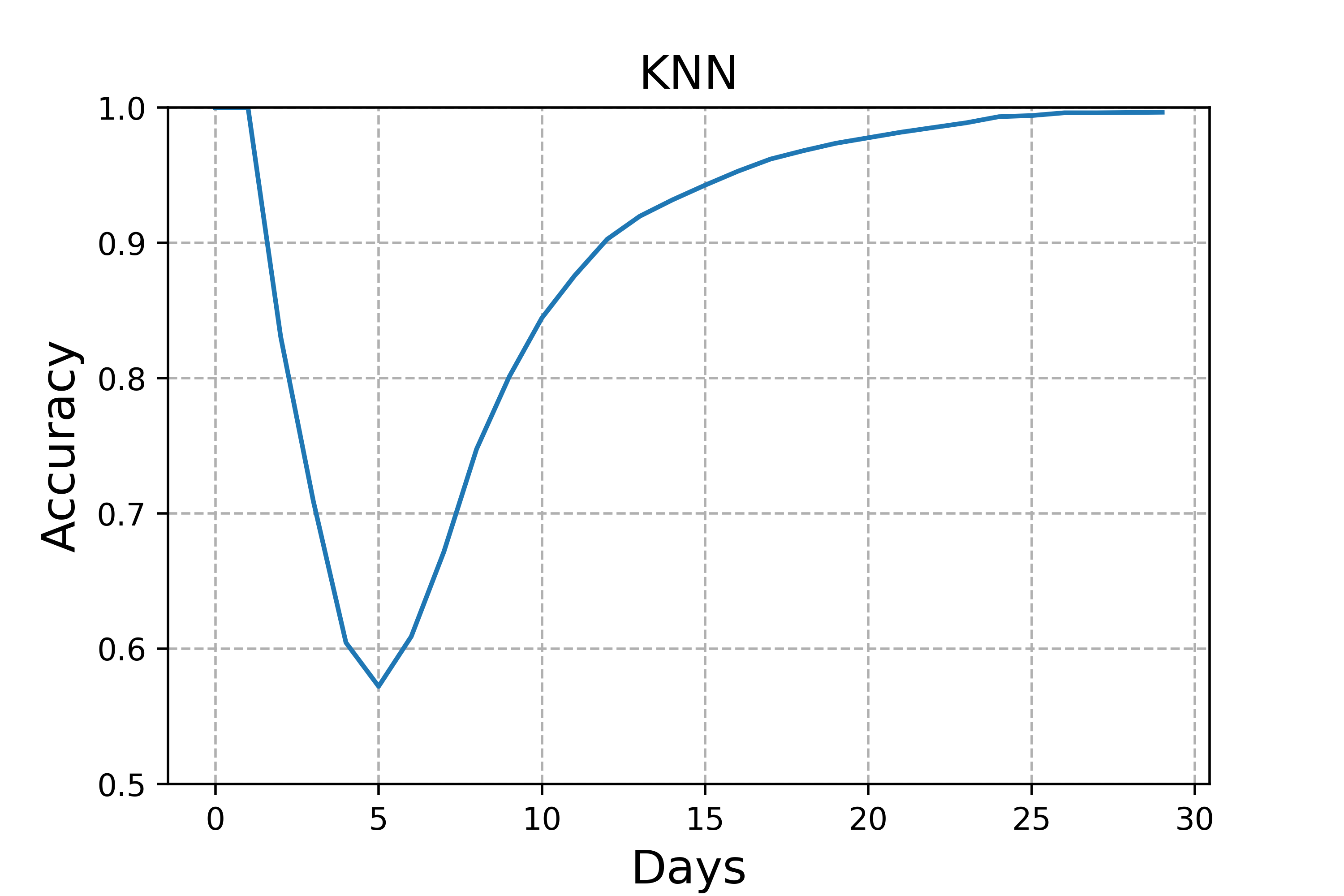}
\includegraphics[width = 6 cm, height = 6 cm]{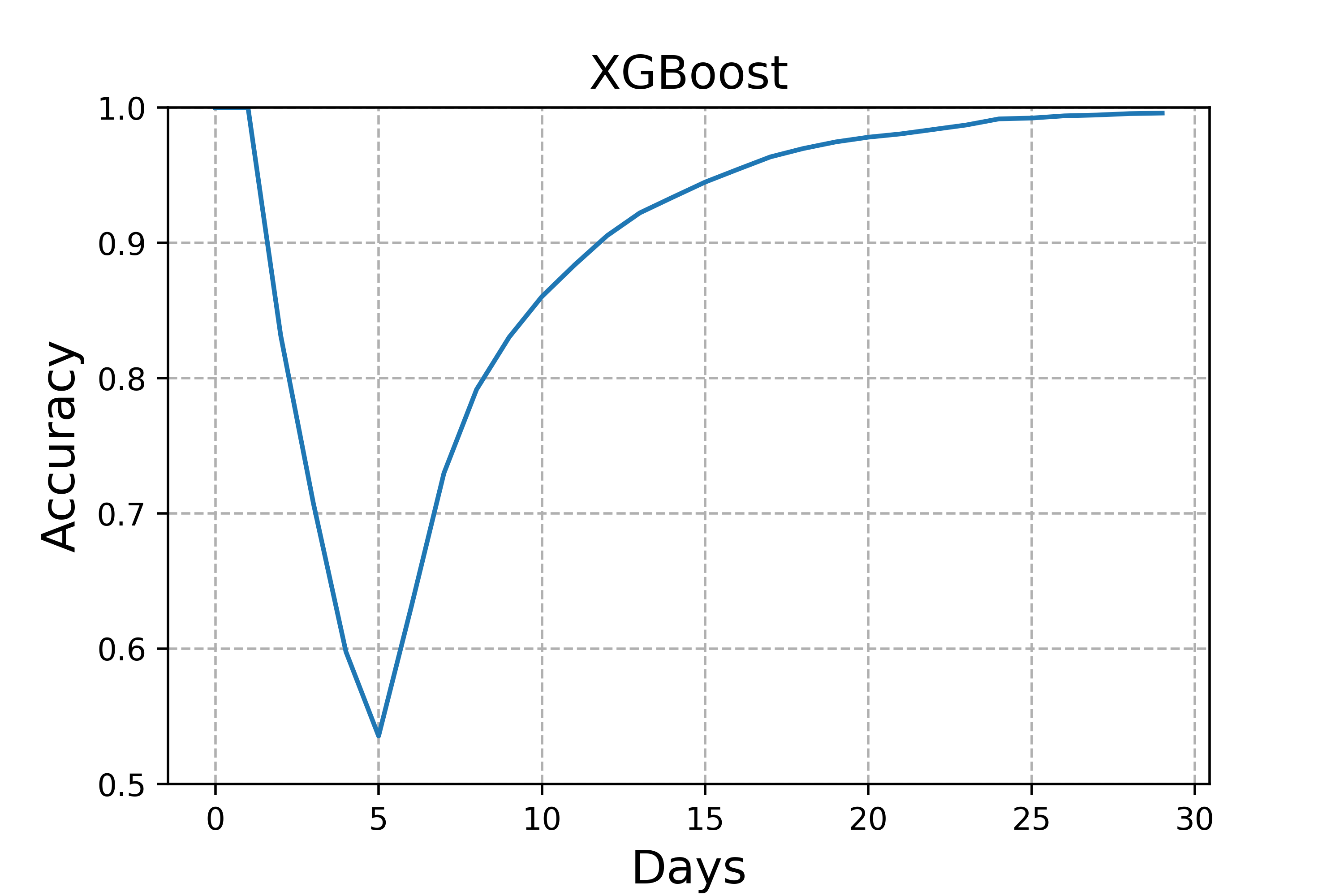}
%\caption{fig1}
\label{regression accuracy}
\end{minipage}
}%
\centering
\caption{regression model result}
\label{regression_result}
\end{figure}

The importance of radius was proved unnecessary as the Trend map shown in Appendix figure \ref{trend_map}. Different Radius presents the similar impact on RMSE value of prediction, thus radius is not a key factor to consider when new models come. All mentioned ML methods were used to predict the fire duration on the test set, fire from 2013 to 2016 on the target area, producing the prediction map for the results from all regression-based ML approaches as shown in figure \ref{regression_result}. Each plot has a predicted value on x label and a true value on y label and the scatter point was used to indicate individual forecasting. The blue scatter line shown on the map represents the real trend of fire duration. Additionally, figure \ref{regression accuracy} give the accuracy plots that provide another measurement to identify the preciseness of models. The feature importance of RF and XGBoost were shown in figure \ref{Feature Importance}. The Outcome Summary can be seen in Table \ref{Result Summary}.

\begin{table}[h]\centering
\caption{Result Summary}
\label{Result Summary}
\begin{tabular}{@{}cccc@{}}
\toprule
Model & RMSE & $R^2 Score$ & Accuracy\\
\midrule
\textbf{Random Forest} & 4.99    & 0.073  & 0.536 \\ 
\textbf{KNN}& 5.26  & -0.031  &  0.572 \\ 
\textbf{XGBoost}& 5.05  &  0.049 & 0.535 \\ 
\textbf{Multi-layer CNN} & 5.05 & 0.051 & 0.629 \\
\textbf{encoder-based CNN} & 5.16 & 0.005 & 0.631\\
\bottomrule
\end{tabular}
\end{table}

\begin{figure}[htbp]
\centering
\subfigure[Prediction Plot]{
\begin{minipage}[ht]{0.45\textwidth}
\centering
    \centering
    \includegraphics[width = 6 cm, height = 6 cm]{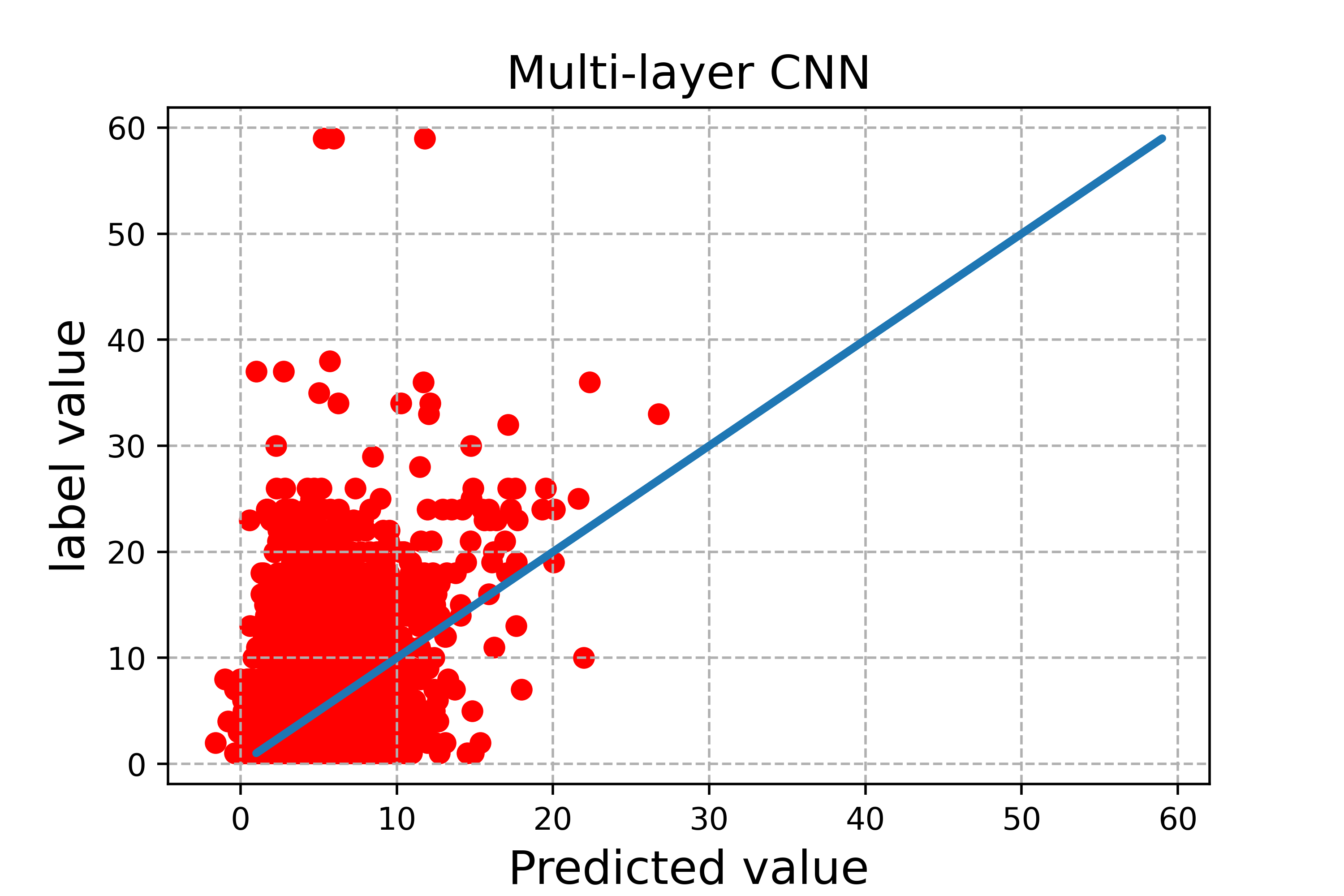}
    \includegraphics[width = 6 cm, height = 6 cm]{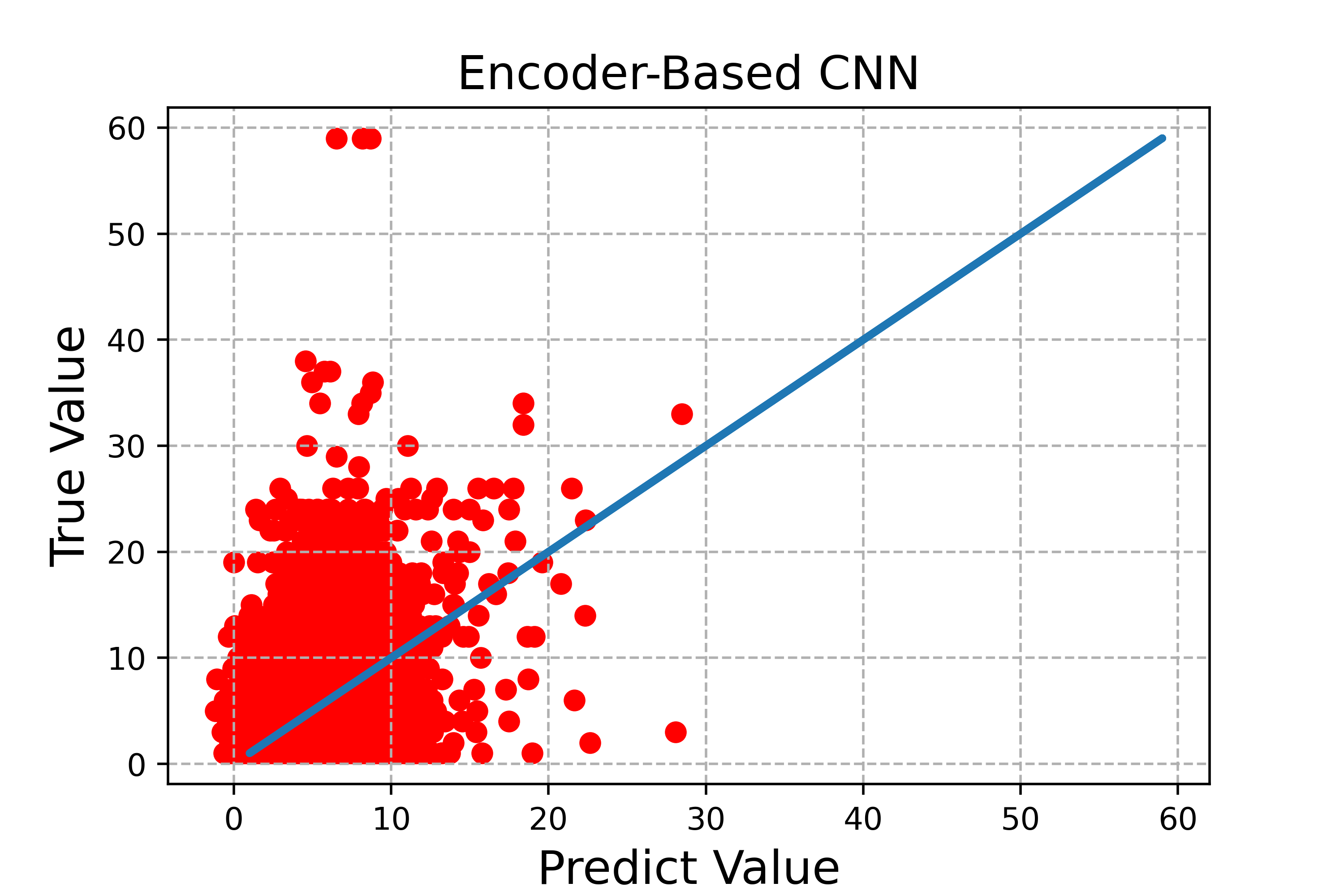}
%\caption{fig1}
\end{minipage}
\label{encoder cnn map}
}%
\subfigure[Accuracy Plot]{
\begin{minipage}[ht]{0.45\textwidth}
\centering

    \includegraphics[width = 6 cm, height = 6 cm]{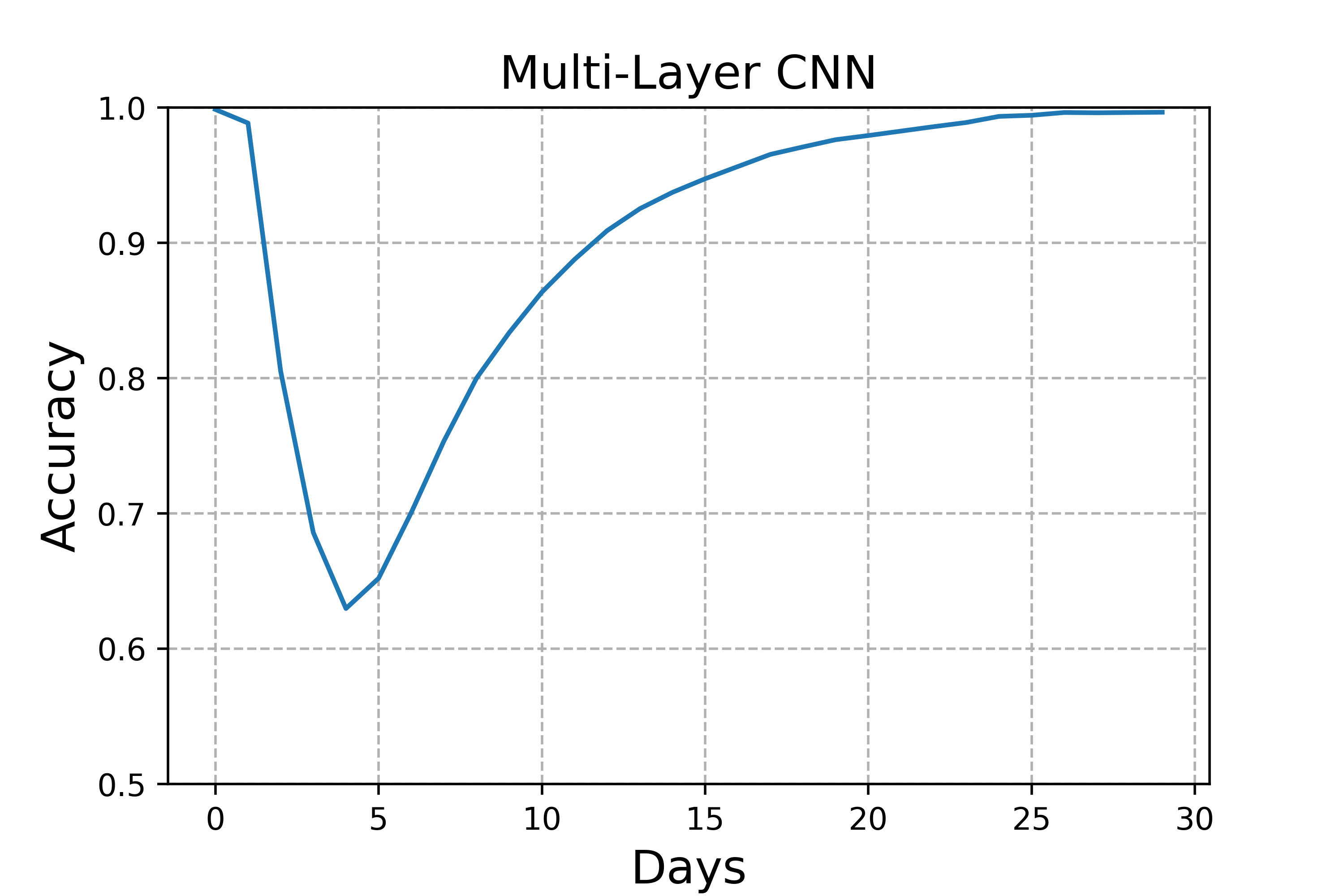}
    \includegraphics[width = 6 cm, height = 6 cm]{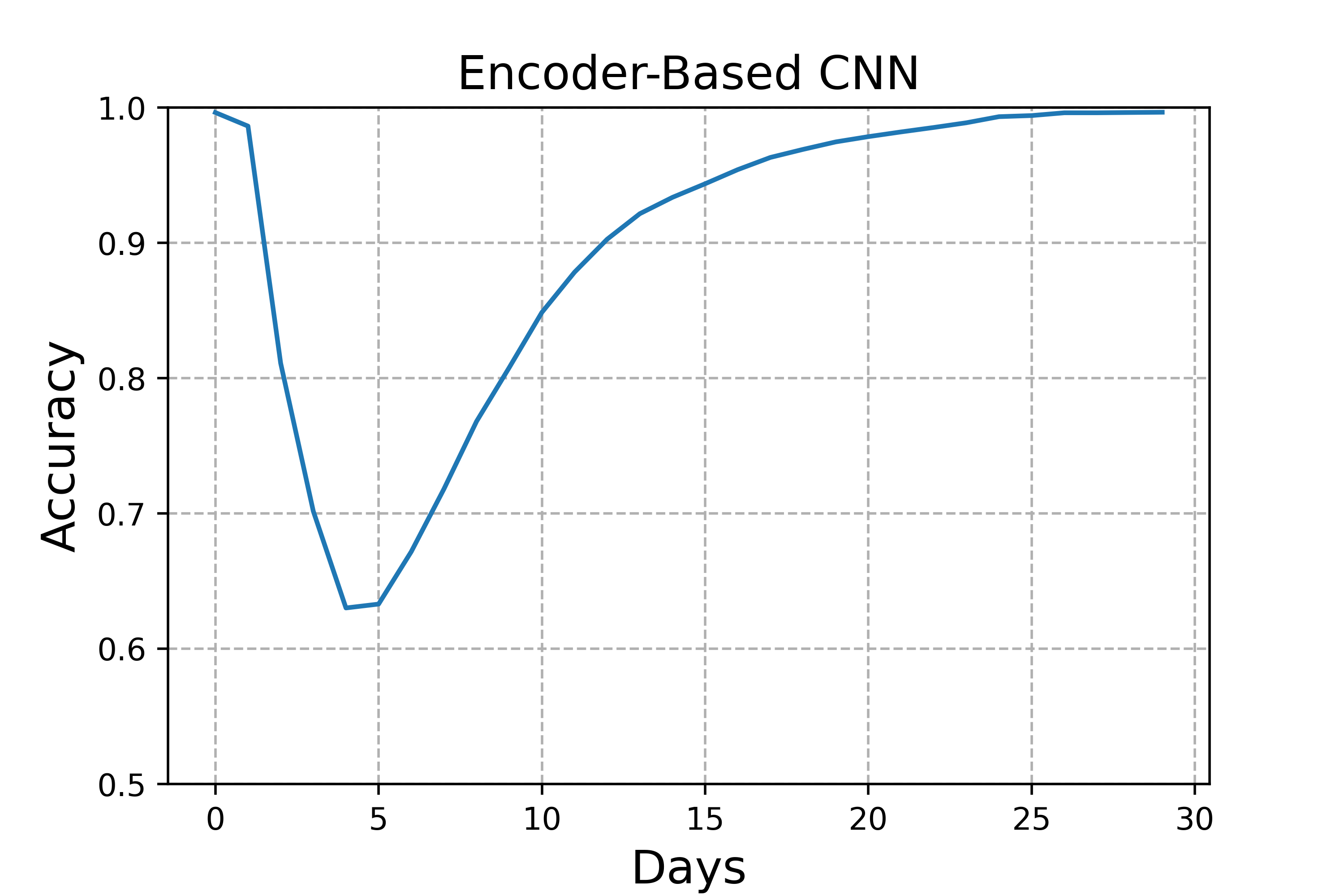}

%\caption{fig1}
\label{CNN}
\end{minipage}
}%
\centering
\caption{CNN model result}
\label{image-based map}
\end{figure}

The result shown in Table \ref{Result Summary} reveals that all applied ML methods presented no big difference on RMSE, although random forest performs the best. Moreover, all the models lack a reasonable $R^2$ score for some reason, probably due to the large number of small fires, which incurs large data bias that is not well quantified by the $R^2$ score. Normally, a $R^2$ score above 0.3 presents a reliable "goodness-of-fit" of the model, but our results give $R^2$ score values all below 0.1, which indicates the inaccuracy of the models. However, some researchers argue that $R^2$ score may be inappropriate to evaluate the performance of a model and state that the accuracy of a model does not only depend on $R^2$ score\citep{rscore}. Thus, the accuracy rate is more valuable than $R^2$ score for referencing the reliability of the model. In the ML regression model, KNN performs the best accuracy, achieving a worst situation around 0.572. RF and XGBoost have nearly the same accuracy rate due to the similarity of these two methods. For image-based models, below two figures \ref{encoder cnn map} and \ref{CNN} give the prediction and accuracy plot of two image-based ML models. It is clear that the image-based model demonstrates a more precise output, which improves by around 14 percent in accuracy. Although the two model presents similar result, they still have distinctiveness. Encoder-Based CNN is more suitable for scenarios in that different sizes of input are involved, multi-layer CNN is applicable to features of the same size. All plots performed an accuracy close to 90\% at days equal to 10 and approaching 100\% at days moving to 30, indicating all the models are more sensitive to long-lasting fire and more precise in predicting large fires.

% figure of regression model result

\newpage
\section{Conclusion}
We developed a machine learning based wildfire prediction model that learns from historical fire data and corresponding landscape maps to achieve fast predictions. The average error of the model can be controlled within one day and has above eighty percent accuracy for fires that last longer than ten days. The models have difficulties in predicting extreme cases, such as the prediction of fires of very long or short duration. Besides, our model cannot achieve a good $R^2$ score as $R^2$ score cannot evaluate the result properly, another statistical coefficient is needed to replace $R^2$ score. This model is not suitable for the case that which small amounts of data and images were provided because our model requires amounts of input to learn a reasonable weight and bias. In addition, the input feature maps may change when applied to new target areas due to the feature importance of landscape features may change in a new region. Future enhancements to the model should focus on improving the regional coverage and generality and the simulation of features not considered in the current situation, utilizing various data augmentation techniques, and applying more encoders for different features. 

\section*{Acknowledgments}
I would like to thank my master project supervisor Dr. S.Cheng for his help in all the problems in the three months, providing model establishing direction, giving crucial advice on model detail, and all the suggestions on writing the report.

\newpage
\bibliographystyle{unsrtnat}
\bibliography{reference}

\begin{thebibliography}{39}
\providecommand{\natexlab}[1]{#1}
\providecommand{\url}[1]{\texttt{#1}}
\expandafter\ifx\csname urlstyle\endcsname\relax
  \providecommand{\doi}[1]{doi: #1}\else
  \providecommand{\doi}{doi: \begingroup \urlstyle{rm}\Url}\fi

\bibitem[ANDELA et~al.(2019)ANDELA, MORTON, GIGLIO, and RANDERSON]{altas_fire}
N.~ANDELA, D.C. MORTON, L.~GIGLIO, and J.T. RANDERSON.
\newblock Global fire atlas with characteristics of individual fires,
  2003-2016, 2019.

\bibitem[McCaffrey(2004)]{1}
Sarah McCaffrey.
\newblock Thinking of wildfire as a natural hazard.
\newblock \emph{Society and Natural Resources}, 17\penalty0 (6):\penalty0
  509--516, 2004.

\bibitem[Bond and Keeley(2005)]{2}
William~J. Bond and Jon~E. Keeley.
\newblock Fire as a global ‘herbivore’: the ecology and evolution of
  flammable ecosystems.
\newblock \emph{Trends in Ecology \& Evolution}, 20\penalty0 (7):\penalty0
  387--394, 2005.
\newblock ISSN 0169-5347.
\newblock \doi{https://doi.org/10.1016/j.tree.2005.04.025}.

\bibitem[Youssouf et~al.(2014)Youssouf, Liousse, Roblou, Assamoi, Salonen,
  Maesano, Banerjee, and Annesi-Maesano]{3}
H.~Youssouf, C.~Liousse, L.~Roblou, E.M. Assamoi, R.O. Salonen, C.~Maesano,
  S.~Banerjee, and I.~Annesi-Maesano.
\newblock Quantifying wildfires exposure for investigating health-related
  effects.
\newblock \emph{Atmospheric Environment}, 97:\penalty0 239--251, 2014.
\newblock ISSN 1352-2310.
\newblock \doi{https://doi.org/10.1016/j.atmosenv.2014.07.041}.

\bibitem[Robinne et~al.(2018)Robinne, Bladon, Miller, Parisien, Mathieu, and
  Flannigan]{4}
François-Nicolas Robinne, Kevin~D. Bladon, Carol Miller, Marc-André Parisien,
  Jérôme Mathieu, and Mike~D. Flannigan.
\newblock A spatial evaluation of global wildfire-water risks to human and
  natural systems.
\newblock \emph{Science of The Total Environment}, 610-611:\penalty0
  1193--1206, 2018.
\newblock ISSN 0048-9697.
\newblock \doi{https://doi.org/10.1016/j.scitotenv.2017.08.112}.

\bibitem[Stetler et~al.(2010)Stetler, Venn, and Calkin]{5}
Kyle~M. Stetler, Tyron~J. Venn, and David~E. Calkin.
\newblock The effects of wildfire and environmental amenities on property
  values in northwest montana, usa.
\newblock \emph{Ecological Economics}, 69\penalty0 (11):\penalty0 2233--2243,
  2010.
\newblock ISSN 0921-8009.
\newblock \doi{https://doi.org/10.1016/j.ecolecon.2010.06.009}.
\newblock Special Section - Payments for Ecosystem Services: From Local to
  Global.

\bibitem[Madigan et~al.(2011)Madigan, Rowe, Angelos, Herthel, Matz, Dinucci,
  and Fletcher]{6}
J.~Madigan, J.~Rowe, J.~Angelos, W.F. Herthel, D.~Matz, M.~Dinucci, and
  V.~Fletcher.
\newblock (a323) wildfire associated burn injury of 1400 sheep in northern
  california: A coordinated mass casualty veterinary response.
\newblock \emph{Prehospital and Disaster Medicine}, 26\penalty0 (S1):\penalty0
  s90–s91, 2011.
\newblock \doi{10.1017/S1049023X11003074}.

\bibitem[Chopard and Droz(1998)]{7}
Bastien Chopard and Michel Droz.
\newblock Cellular automata.
\newblock \emph{Modelling of Physical}, 1998.

\bibitem[Anderson and Wendt(1995)]{8}
John~David Anderson and John Wendt.
\newblock \emph{Computational fluid dynamics}, volume 206.
\newblock Springer, 1995.

\bibitem[Rothermel(1972)]{rothermel1972mathematical}
Richard~C Rothermel.
\newblock \emph{A mathematical model for predicting fire spread in wildland
  fuels}, volume 115.
\newblock Intermountain Forest \& Range Experiment Station, Forest Service,
  US~…, 1972.

\bibitem[Alexandridis et~al.(2008)Alexandridis, Vakalis, Siettos, and
  Bafas]{cellular_automate}
Alex Alexandridis, D~Vakalis, Constantinos~I Siettos, and George~V Bafas.
\newblock A cellular automata model for forest fire spread prediction: The case
  of the wildfire that swept through spetses island in 1990.
\newblock \emph{Applied Mathematics and Computation}, 204\penalty0
  (1):\penalty0 191--201, 2008.

\bibitem[Fradkov(2020)]{mlhistory}
Alexander~L Fradkov.
\newblock Early history of machine learning.
\newblock \emph{IFAC-PapersOnLine}, 53\penalty0 (2):\penalty0 1385--1390, 2020.

\bibitem[Cheng et~al.(2022{\natexlab{a}})Cheng, Prentice, Huang, Jin, Guo, and
  Arcucci]{cheng2022data}
Sibo Cheng, I~Colin Prentice, Yuhan Huang, Yufang Jin, Yi-Ke Guo, and Rossella
  Arcucci.
\newblock Data-driven surrogate model with latent data assimilation:
  Application to wildfire forecasting.
\newblock \emph{Journal of Computational Physics}, 464:\penalty0 111302,
  2022{\natexlab{a}}.

\bibitem[Cheng et~al.(2022{\natexlab{b}})Cheng, Jin, Harrison,
  Quilodr{\'a}n-Casas, Prentice, Guo, and Arcucci]{cheng2022parameter}
Sibo Cheng, Yufang Jin, Sandy~P Harrison, C{\'e}sar Quilodr{\'a}n-Casas,
  Iain~Colin Prentice, Yi-Ke Guo, and Rossella Arcucci.
\newblock Parameter flexible wildfire prediction using machine learning
  techniques: Forward and inverse modelling.
\newblock \emph{Remote Sensing}, 14\penalty0 (13):\penalty0 3228,
  2022{\natexlab{b}}.

\bibitem[Zhong et~al.(2023)Zhong, Cheng, Kasoar, and Arcucci]{zhong2023reduced}
Caili Zhong, Sibo Cheng, Matthew Kasoar, and Rossella Arcucci.
\newblock Reduced-order digital twin and latent data assimilation for global
  wildfire prediction.
\newblock \emph{Natural Hazards and Earth System Sciences}, 23\penalty0
  (5):\penalty0 1755--1768, 2023.

\bibitem[Cheng et~al.(2023{\natexlab{a}})Cheng, Guo, and
  Arcucci]{cheng2023generative}
Sibo Cheng, Yike Guo, and Rossella Arcucci.
\newblock A generative model for surrogates of spatial-temporal wildfire
  nowcasting.
\newblock \emph{IEEE Transactions on Emerging Topics in Computational
  Intelligence}, 2023{\natexlab{a}}.

\bibitem[Rodrigues and {de la Riva}(2014)]{human-caused_wildfire}
Marcos Rodrigues and Juan {de la Riva}.
\newblock An insight into machine-learning algorithms to model human-caused
  wildfire occurrence.
\newblock \emph{Environmental Modelling \& Software}, 57:\penalty0 192--201,
  2014.
\newblock ISSN 1364-8152.
\newblock \doi{https://doi.org/10.1016/j.envsoft.2014.03.003}.

\bibitem[Lever et~al.(2023)Lever, Cheng, and Arcucci]{lever2023social}
Jake Lever, Sibo Cheng, and Rossella Arcucci.
\newblock Social \& physics based data driven methods for wildfire prediction.
\newblock Technical report, Copernicus Meetings, 2023.

\bibitem[Parisien et~al.(2012)Parisien, Snetsinger, Greenberg, Nelson,
  Schoennagel, Dobrowski, and Moritz]{probabilityofwildfire}
Marc-Andr{\'e} Parisien, Susan Snetsinger, Jonathan~A Greenberg, Cara~R Nelson,
  Tania Schoennagel, Solomon~Z Dobrowski, and Max~A Moritz.
\newblock Spatial variability in wildfire probability across the western united
  states.
\newblock \emph{International Journal of Wildland Fire}, 21\penalty0
  (4):\penalty0 313--327, 2012.

\bibitem[Breiman(2001)]{random_forest}
Leo Breiman.
\newblock Statistical modeling: The two cultures (with comments and a rejoinder
  by the author).
\newblock \emph{Statistical science}, 16\penalty0 (3):\penalty0 199--231, 2001.

\bibitem[Altman(1992)]{kNN}
N.~S. Altman.
\newblock An introduction to kernel and nearest-neighbor nonparametric
  regression.
\newblock \emph{The American Statistician}, 46\penalty0 (3):\penalty0 175--185,
  1992.
\newblock \doi{10.1080/00031305.1992.10475879}.

\bibitem[Gong et~al.(2022)Gong, Cheng, Chen, Li, Quilodr{\'a}n-Casas, Xiao, and
  Arcucci]{gong2022efficient}
Helin Gong, Sibo Cheng, Zhang Chen, Qing Li, C{\'e}sar Quilodr{\'a}n-Casas,
  Dunhui Xiao, and Rossella Arcucci.
\newblock An efficient digital twin based on machine learning svd autoencoder
  and generalised latent assimilation for nuclear reactor physics.
\newblock \emph{Annals of nuclear energy}, 179:\penalty0 109431, 2022.

\bibitem[Chen and Guestrin(2016)]{chen2016xgboost}
Tianqi Chen and Carlos Guestrin.
\newblock Xgboost: A scalable tree boosting system.
\newblock In \emph{Proceedings of the 22nd acm sigkdd international conference
  on knowledge discovery and data mining}, pages 785--794, 2016.

\bibitem[Cheng et~al.(2023{\natexlab{b}})Cheng, Chen, Anastasiou, Angeli,
  Matar, Guo, Pain, and Arcucci]{cheng2023generalised}
Sibo Cheng, Jianhua Chen, Charitos Anastasiou, Panagiota Angeli, Omar~K Matar,
  Yi-Ke Guo, Christopher~C Pain, and Rossella Arcucci.
\newblock Generalised latent assimilation in heterogeneous reduced spaces with
  machine learning surrogate models.
\newblock \emph{Journal of Scientific Computing}, 94\penalty0 (1):\penalty0 11,
  2023{\natexlab{b}}.

\bibitem[Kandakji et~al.(2020)Kandakji, Gill, and Lee]{geofeature}
Tarek Kandakji, Thomas~E. Gill, and Jeffrey~A. Lee.
\newblock Identifying and characterizing dust point sources in the southwestern
  united states using remote sensing and gis.
\newblock \emph{Geomorphology}, 353:\penalty0 107019, 2020.
\newblock ISSN 0169-555X.
\newblock \doi{https://doi.org/10.1016/j.geomorph.2019.107019}.

\bibitem[Dennison et~al.(2014)Dennison, Brewer, Arnold, and Moritz]{firetrend}
Philip~E Dennison, Simon~C Brewer, James~D Arnold, and Max~A Moritz.
\newblock Large wildfire trends in the western united states, 1984--2011.
\newblock \emph{Geophysical Research Letters}, 41\penalty0 (8):\penalty0
  2928--2933, 2014.

\bibitem[Boesch et~al.(2000)Boesch, Field, and Scavia]{ecoloss}
Donald~F Boesch, John~C Field, and Donald Scavia.
\newblock \emph{The potential consequences of climate variability and change on
  coastal areas and marine resources: Report of the Coastal Areas and Marine
  Resources Sector Team, US National Assessment of the Potential Consequences
  of Climate Variability and Change, US Global Change Research Program}.
\newblock Number~21. US Department of Commerce, National Oceanic and
  Atmospheric Administration~…, 2000.

\bibitem[Radke et~al.(2019)Radke, Hessler, and Ellsworth]{duration}
David Radke, Anna Hessler, and Dan Ellsworth.
\newblock Firecast: Leveraging deep learning to predict wildfire spread.
\newblock In \emph{IJCAI}, pages 4575--4581, 2019.

\bibitem[Spawn et~al.(2020)Spawn, Sullivan, Lark, and Gibbs]{biomass}
Seth~A. Spawn, Clare~C. Sullivan, Tyler~J. Lark, and Holly~K. Gibbs.
\newblock Harmonized global maps of above and belowground biomass carbon
  density in the year 2010.
\newblock \emph{Scientific Data}, 7\penalty0 (1):\penalty0 112, Apr 2020.
\newblock ISSN 2052-4463.
\newblock \doi{10.1038/s41597-020-0444-4}.

\bibitem[Theobald et~al.(2015)Theobald, Harrison-Atlas, Monahan, and
  Albano]{slope}
David~M. Theobald, Dylan Harrison-Atlas, William~B. Monahan, and Christine~M.
  Albano.
\newblock Ecologically-relevant maps of landforms and physiographic diversity
  for climate adaptation planning.
\newblock \emph{PLOS ONE}, 10:\penalty0 1--17, 12 2015.
\newblock \doi{10.1371/journal.pone.0143619}.

\bibitem[Buchhorn et~al.(2020)Buchhorn, Lesiv, Tsendbazar, Herold, Bertels, and
  Smets]{landcover}
Marcel Buchhorn, Myroslava Lesiv, Nandin-Erdene Tsendbazar, Martin Herold, Luc
  Bertels, and Bruno Smets.
\newblock Copernicus global land cover layers—collection 2.
\newblock \emph{Remote Sensing}, 12\penalty0 (6), 2020.
\newblock ISSN 2072-4292.
\newblock \doi{10.3390/rs12061044}.

\bibitem[Hersbach et~al.(2018)Hersbach, Bell, Berrisford, Biavati, Hor{\'a}nyi,
  Mu{\~n}oz~Sabater, Nicolas, Peubey, Radu, Rozum, et~al.]{precipitation}
H~Hersbach, B~Bell, P~Berrisford, G~Biavati, A~Hor{\'a}nyi,
  J~Mu{\~n}oz~Sabater, J~Nicolas, C~Peubey, R~Radu, I~Rozum, et~al.
\newblock Era5 hourly data on single levels from 1979 to present.
\newblock \emph{Copernicus Climate Change Service (C3S) Climate Data Store
  (CDS)}, 10, 2018.

\bibitem[Jayalakshmi and Santhakumaran(2011)]{minmax}
T~Jayalakshmi and A~Santhakumaran.
\newblock Statistical normalization and back propagation for classification.
\newblock \emph{International Journal of Computer Theory and Engineering},
  3\penalty0 (1):\penalty0 1793--8201, 2011.

\bibitem[Ke et~al.(2017)Ke, Meng, Finley, Wang, Chen, Ma, Ye, and
  Liu]{ke2017lightgbm}
Guolin Ke, Qi~Meng, Thomas Finley, Taifeng Wang, Wei Chen, Weidong Ma, Qiwei
  Ye, and Tie-Yan Liu.
\newblock Lightgbm: A highly efficient gradient boosting decision tree.
\newblock \emph{Advances in neural information processing systems}, 30, 2017.

\bibitem[Smith(1981)]{bi-linear}
P.R. Smith.
\newblock Bilinear interpolation of digital images.
\newblock \emph{Ultramicroscopy}, 6\penalty0 (2):\penalty0 201--204, 1981.
\newblock ISSN 0304-3991.
\newblock \doi{https://doi.org/10.1016/0304-3991(81)90061-9}.

\bibitem[Albawi et~al.(2017)Albawi, Mohammed, and Al-Zawi]{CNN}
Saad Albawi, Tareq~Abed Mohammed, and Saad Al-Zawi.
\newblock Understanding of a convolutional neural network.
\newblock In \emph{2017 international conference on engineering and technology
  (ICET)}, pages 1--6. Ieee, 2017.

\bibitem[Mastromichalakis(2020)]{leakyrelu}
Stamatis Mastromichalakis.
\newblock Alrelu: A different approach on leaky relu activation function to
  improve neural networks performance.
\newblock \emph{arXiv preprint arXiv:2012.07564}, 2020.

\bibitem[Ng et~al.(2011)]{sparseautoencoder}
Andrew Ng et~al.
\newblock Sparse autoencoder.
\newblock \emph{CS294A Lecture notes}, 72\penalty0 (2011):\penalty0 1--19,
  2011.

\bibitem[Onyutha(2020)]{rscore}
Charles Onyutha.
\newblock From r-squared to coefficient of model accuracy for assessing"
  goodness-of-fits".
\newblock \emph{Geoscientific Model Development Discussions}, pages 1--25,
  2020.

\end{thebibliography}
% For bibLaTeX users:
% \printbibliography

\newpage
\appendix
\section{Appendix}
\section*{Main notations and acronyms}
\begin{table*}[ht!]
    \centering
    \begin{tabular}{ p{3.5cm} p{9cm}}
$^r{\mathbf{X}}^{tc}_{z}$ & Image of tree cover $tc$ with fire index $z$ in Radius $r$  \\
$^r{\mathbf{X}}^{gc}_{z}$ & Image of grass cover $gc$ with fire number $z$ in Radius $r$  \\
$^r{\mathbf{X}}^{slope}_{z}$ & Image of slope $slope$ with fire number $z$ in Radius $r$  \\
$^r{\mathbf{D}}_z^{model}$ & Predicted Duration of model with fire number $z$ in Radius $r$\\
$CNN$ & Convolution Neural Network \\
$RF$ & Random Forest \\
$KNN$ & K-Nearest Neighbors \\
$XGBoost$ &  eXtreme Gradient Boosting \\
$CFD$ & Computational fluid dynamics \\
    \end{tabular}
\end{table*}

\section{Supplyment Figures}
\begin{figure}[h]
    \centering
    \includegraphics[width = 0.8\textwidth]{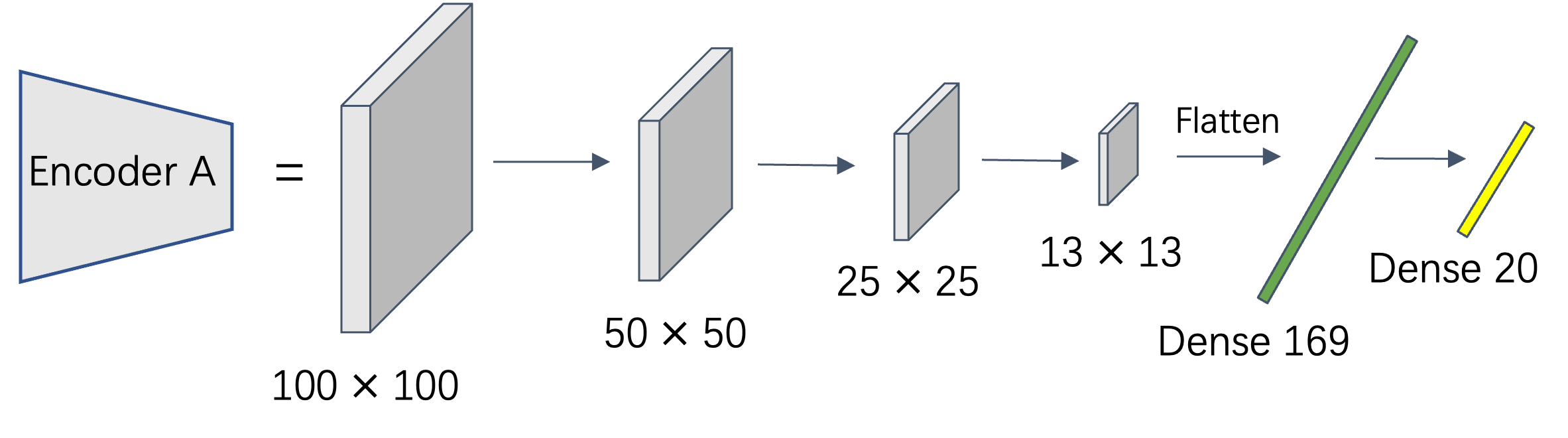}
    \includegraphics[width = 0.8\textwidth]{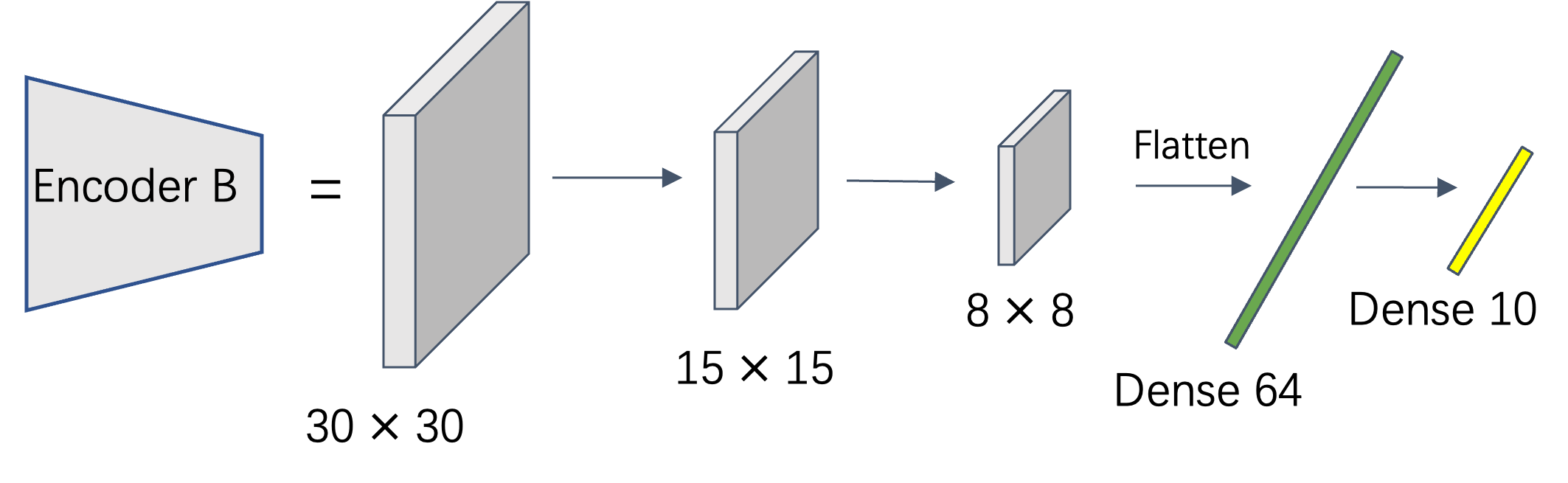}
    \includegraphics[width = 0.8\textwidth]{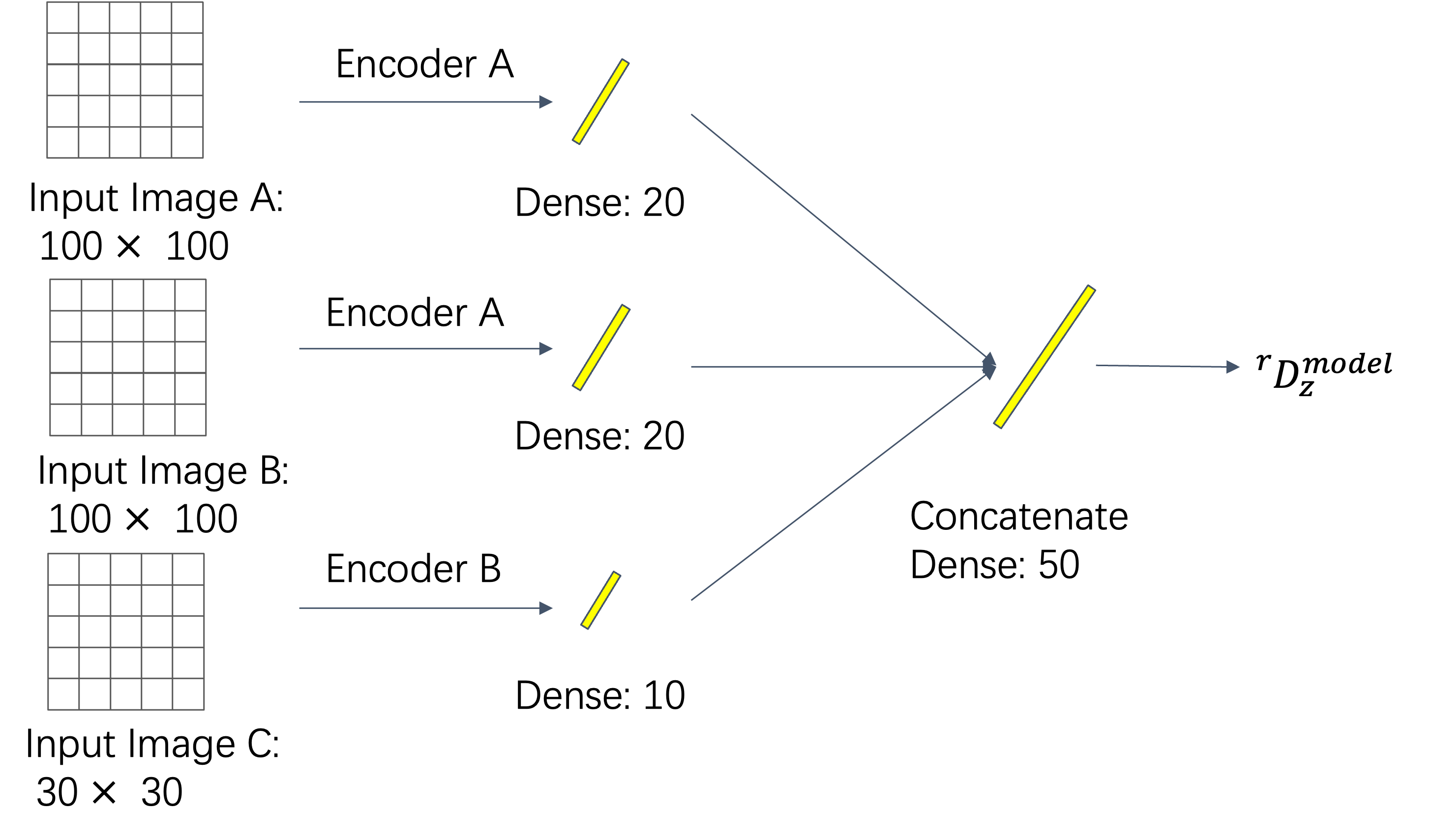}
    \caption{Encoder Based CNN Structure}
    \label{encoder_based_cnn}
\end{figure}

\begin{figure}[h]
    \centering
    \includegraphics[width = 10cm, height = 10cm]{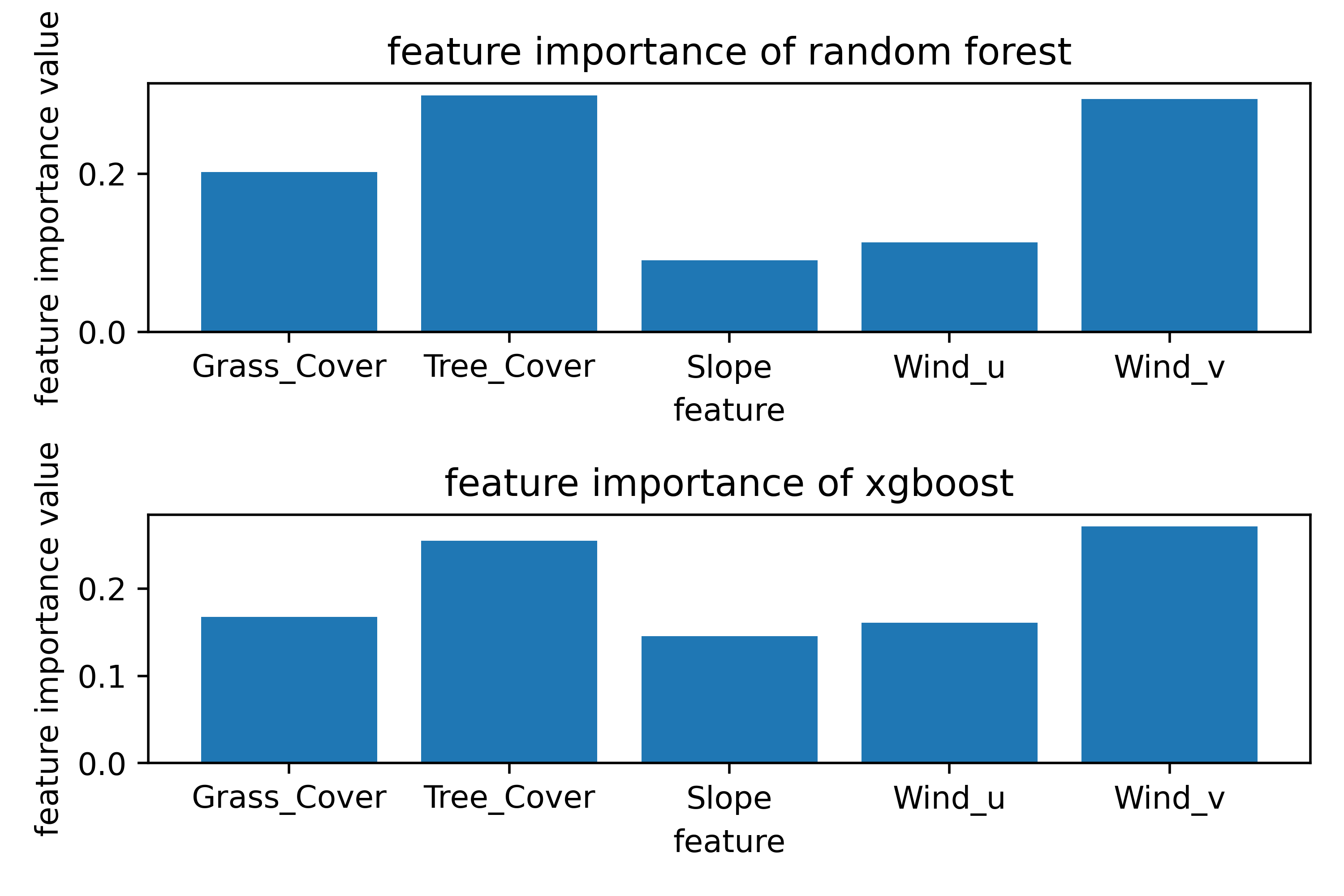}
     \caption{Feature Importance Plot}
\label{Feature Importance}
\end{figure}

\begin{figure}[h]
    \centering
    \includegraphics[width = 10cm, height = 8cm]{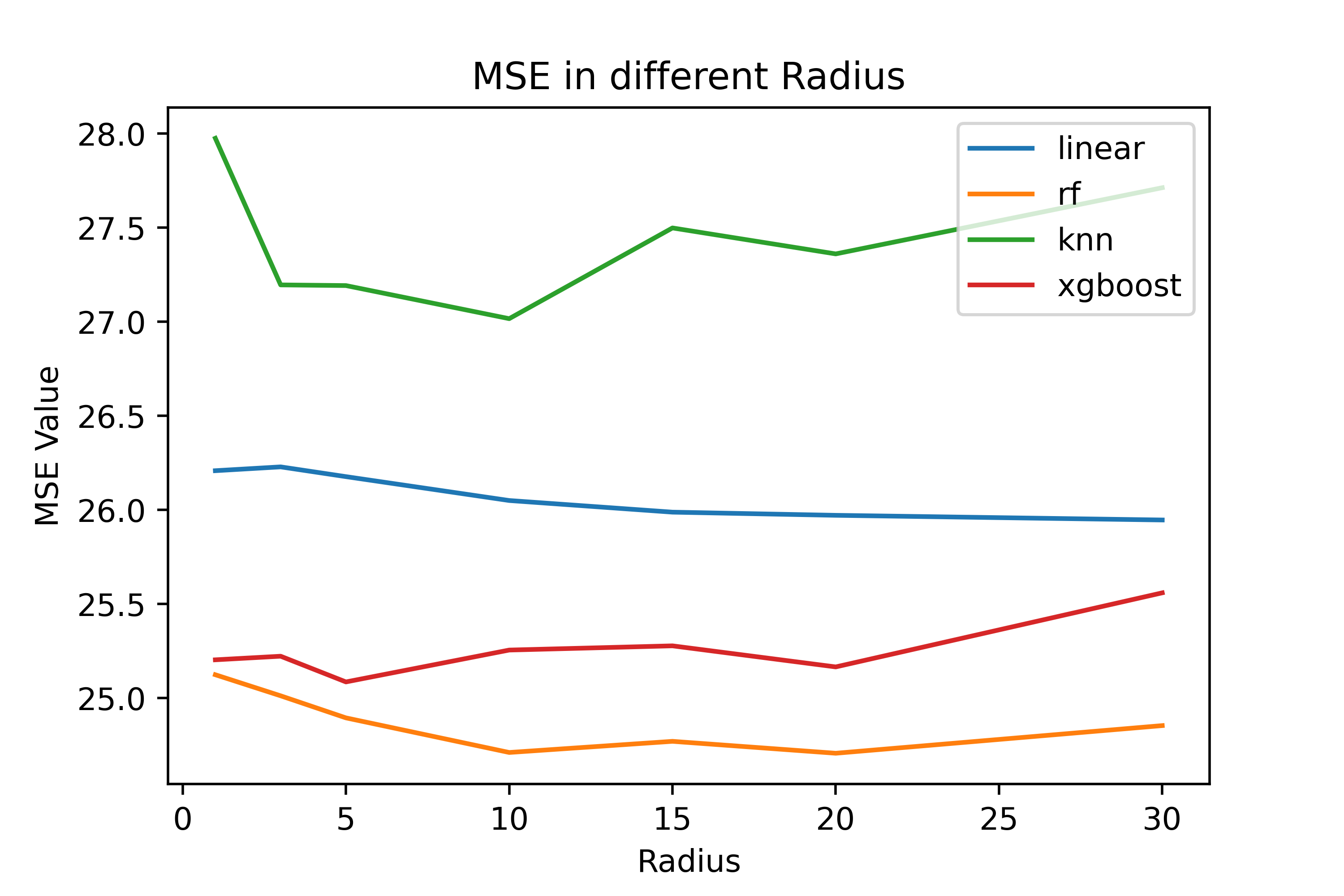}
    \caption{Trend map}
    \label{trend_map}
\end{figure}
\end{document}